\documentclass[journal]{IEEEtran}

\usepackage{amsmath}
\usepackage{cases}
\usepackage{subcaption}
\usepackage{subfig}
\usepackage{float}
\usepackage{graphicx}
\usepackage{algorithm2e}

\usepackage{enumitem}
\usepackage{cite}

\begin{document}

\title{Mapping Generative Models onto a\\Network of Digital Spiking Neurons}

\author{Bruno U. Pedroni, Srinjoy Das, John V. Arthur, Paul A. Merolla, Bryan L. Jackson, Dharmendra S. Modha, Kenneth Kreutz-Delgado, and Gert Cauwenberghs

\thanks{Manuscript received October 9, 2015. This work was supported by the DARPA SyNAPSE Program, the National Science Foundation Emerging Frontiers in Research Innovation ENG-1137279 and Expeditions in Computing CCF-1317407, the Office of Naval Research (ONR MURI 14-13-1-0205), and CNPq Brazil (CsF 201174/2012-0).}

\thanks{B.U. Pedroni and G. Cauwenberghs are with the Department of Bioengineering and the Institute for Neural Computation, UC San Diego, La Jolla, CA 92093 USA (e-mails: bpedroni@eng.ucsd.edu, gert@ucsd.edu).}

\thanks{S. Das and K. Kreutz-Delgado are with the Electrical and Computer Engineering Department, UC San Diego, La Jolla, CA 92093 USA (e-mails: s2das@ucsd.edu, kreutz@eng.ucsd.edu).}

\thanks{J.V. Arthur, P.A. Merolla, B.L. Jackson, and D.S. Modha are with IBM Research - Almaden, San Jose, CA 95120 USA (e-mails: arthurjo, pameroll, bryanlj, dmodha@us.ibm.com).}}

%
%

\markboth{This manuscript has been submitted to IEEE TBioCAS for revision in October 2015}%
{}

%



\maketitle

\begin{abstract}
Stochastic neural networks such as Restricted Boltzmann Machines (RBMs) have been successfully used in applications ranging from speech recognition to image classification. Inference and learning in these algorithms use a Markov Chain Monte Carlo procedure called Gibbs sampling, where a logistic function forms the kernel of this sampler. On the other side of the spectrum, neuromorphic systems have shown great promise for low-power and parallelized cognitive computing, but lack well-suited applications and automation procedures. In this work, we propose a systematic method for bridging the RBM algorithm and digital neuromorphic systems, with a generative pattern completion task as proof of concept. For this, we first propose a method of producing the Gibbs sampler using bio-inspired digital noisy integrate-and-fire neurons. Next, we describe the process of mapping generative RBMs trained offline onto the IBM TrueNorth neurosynaptic processor -- a low-power digital neuromorphic VLSI substrate. Mapping these algorithms onto neuromorphic hardware presents unique challenges in network connectivity and weight and bias quantization, which, in turn, require architectural and design strategies for the physical realization. Generative performance metrics are analyzed to validate the neuromorphic requirements and to best select the neuron parameters for the model. Lastly, we describe a design automation procedure which achieves optimal resource usage, accounting for the novel hardware adaptations. This work represents the first implementation of generative RBM inference on a neuromorphic VLSI substrate.
\end{abstract}


%
\IEEEpeerreviewmaketitle

\section{Introduction}

\begin{figure}[hbtp]
	\centering
	\includegraphics[width=0.49\textwidth]{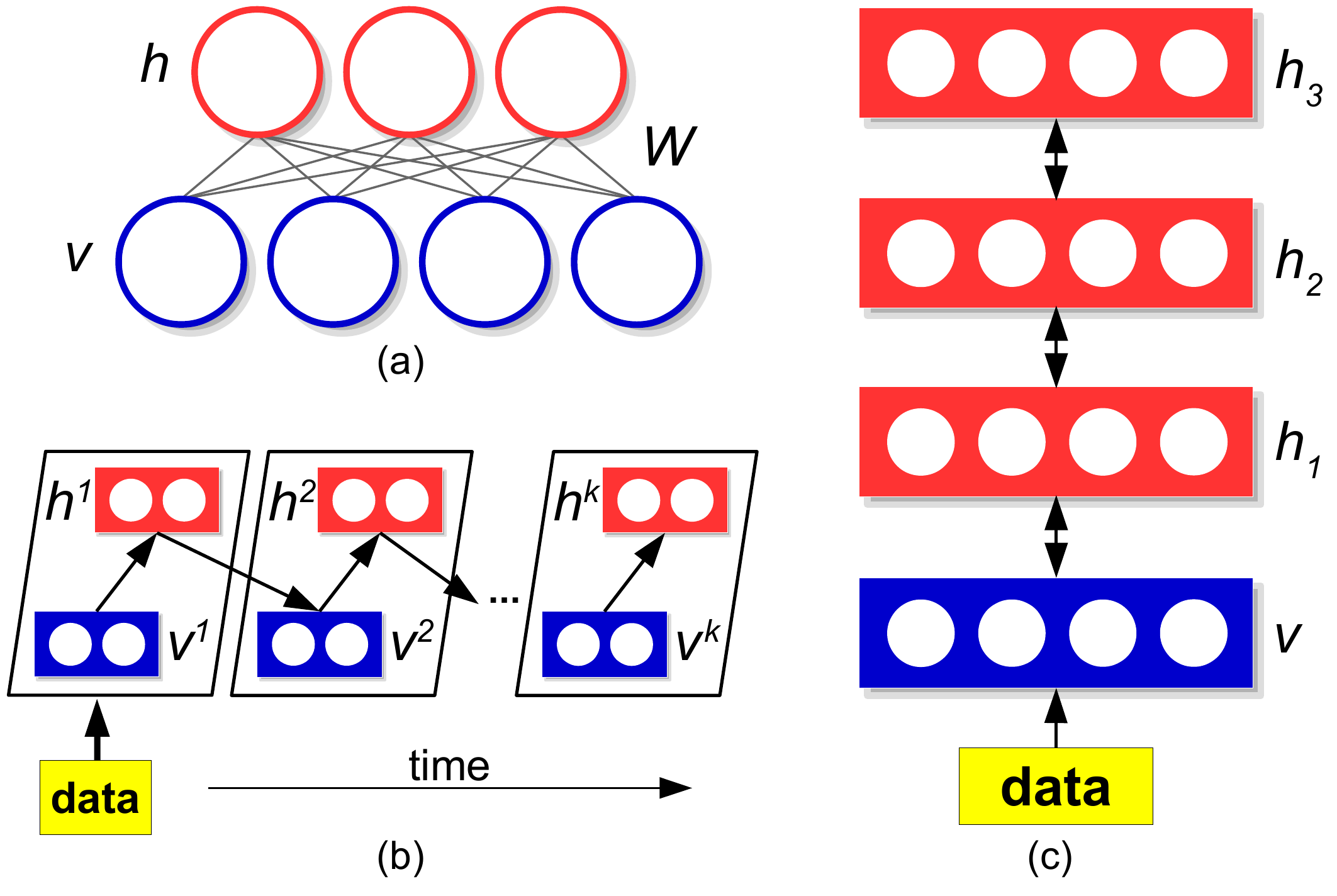}
	\caption{RBM and DBN representations. (a) An RBM formed by 4 visible and 3 hidden units. (b) Gibbs sampling procedure in an RBM. (c) A DBN formed by stacking RBMs.}
	\label{fig:rbm}
\end{figure}

\IEEEPARstart{D}{eep} Learning algorithms such as Restricted Boltzmann Machines (RBMs) and Deep Belief Networks (DBNs) have been successfully used in a wide range of cognitive computing applications such as image classification \cite{larochelle2008classification}, speech recognition \cite{lee2009unsupervised,dahl2012context}, and motion synthesis \cite{taylor2007modeling}. Additionally, these algorithms have been explored as possible solutions for Brain-Computer Interfaces (BCI) and electroencephalography (EEG) data feature learning and classification \cite{wulsin2011modeling,an2014deep}. RBMs are generative learning algorithms and are particularly useful in extracting features from unlabeled data (i.e. unsupervised learning) \cite{hinton2006reducing}. Structurally, an RBM is a stochastic neural network composed of 2 layers of neuron-like units: a layer of visible units $v$ which are driven by the real-world data of interest and a layer of hidden units $h$ which form connections to these visible units. There are no interconnections within a layer and the weights of connections between layers are symmetric. Fig.~\ref{fig:rbm}a exemplifies an RBM with 4 visible and 3 hidden units. 

The RBM defines a joint probability over the input data and hidden variables specified by the Boltzmann distribution \cite{haykin08neural}:

\begin{equation} \label{eq:energy} p(\textbf{v,h}) = \frac {e^{-E(\textbf{v,h})}}{\sum_{\textbf{v,h}} e^{-E(\textbf{v,h})}}, \end{equation}
\vspace{+3pt}where $E(\textbf{v,h}) = -\textbf{v}^{T}\textbf{W}\textbf{h} - \textbf{b}_{v}^{T}\textbf{v} - \textbf{b}_{h}^{T}\textbf{h}$. Here $p$ denotes the Boltzmann probability distribution and $E$ is a function (also known as the ``energy function'') of $\textbf{v}$ and $\textbf{h}$, where $\textbf{v}$ denotes the binary state (0 or 1) of the visible units and $\textbf{h}$ represents the binary state of the hidden units. The weight between visible and hidden units is represented by $\textbf{W}$, while $b_v$ and $b_h$ represent the biases of $v$ and $h$, respectively. The denominator is the sum of all possible states of visible and hidden units, also known as the partition function.


Inference in an RBM can be performed using a Markov Chain Monte Carlo (MCMC) procedure called Gibbs sampling, where each unit in any given layer is sampled conditioning on its total input from units in the other layer. Fig.~\ref{fig:rbm}b illustrates $k$ steps of MCMC performed in an RBM. The Gibbs sampling rule in binary RBMs is defined by the logistic function,
\begin{equation}\label{eq:sigmoid} \sigma(x)= 1/(1+e^{-x}), \end{equation}
with the probability of activation of unit $i$ as defined by \cite{haykin08neural}
\begin{equation}\label{eq:logistic} p(x_{i} = 1 | x_{j}) = \sigma(\sum_{j}w_{ij}x_{j} + b_{i}), \end{equation}
where $w_{ij}$ is the weight from unit $j$ to unit $i$ for all $j \notin layer(i)$, and $b_i$ is the bias of unit $i$. DBNs are formed by stacking layers of RBMs (Fig.~\ref{fig:rbm}c) and it has been shown that inference in a DBN can be done in a successive layer-by-layer manner on each RBM \cite{hinton2006fast}. RBMs and DBNs can be used with labeled data for classification tasks either as feature extractors to an external classifier or as completely self-contained discriminative machine learning frameworks \cite{larochelle2008classification}. However, most of the data in the real world is unlabeled and, in such situations, RBMs and DBNs can be used to perform generative inference tasks. Applications of inference in such unsupervised frameworks include, for example, restoration of incomplete or occluded images and prediction of motion sequences.

Currently, inference tasks using RBMs and DBNs are overwhelmingly realized in software, which are typically run on high performance CPUs (Central Processing Unit) and GPUs (Graphical Processing Unit). For ultra low-power, real-time realizations of these algorithms, such as in mobile devices, the solution tends to be sending information to the cloud for processing. However, this demands, in many cases, reliable communication between client and server, along with large amounts of transmitted data. In this context, the Neuromorphic Computing paradigm is a more suitable solution in terms of low-power client-side processing. Neuromorphic VLSI (Very Large Scale Integrated Circuit) systems \cite{Indiveri2006,khan2008spinnaker,mitra2009real,basu2010neural,zamarreno2013multicasting,benjamin2014neurogrid,merolla2014million,mayr2014biological}, inspired by biological neural architectures and functions, have been realized with analog, digital, and mixed-signal circuit elements. Such systems typically compute in a massively parallel fashion and communicate asynchronously using spikes. The principal benefit of this architecture, which stands in contrast to the traditional von Neumann computing paradigm, is extremely energy efficient computation in a highly concurrent fashion. Algorithms which demand large matrix multiplications, such as RBMs and DBNs, benefit greatly in terms of computation (and, consequently, power) when implemented in spike-based systems, mainly because multiplications by zero are avoided (i.e. absence of spikes does not generate computation). Therefore, arrays of spiking neurons realized on neuromorphic VLSI are ideal for classification, generation and other inference tasks in the context of real-world high dimensional data.

The goal of our work is to develop a modular architecture in a systematic fashion to form a foundation for building neural networks, such as RBMs and DBNs, on substrates of digital spiking neurons. As a proof of concept of our design approach, we implement a pre-trained (i.e. trained offline) generative RBM for pattern completion on the TrueNorth digital neuromorphic VLSI device using the MNIST handwritten digit images dataset.

The remainder of this paper is divided in the following manner: Section \ref{sec:markov} describes the Markov chain analysis of the digital neural sampler; Section \ref{sec:truenorth} describes the TrueNorth system and the challenges in implementing Deep Learning algorithms, along with the necessary steps for mapping the RBM algorithm onto digital spiking neuromorphic hardware; Section \ref{sec:quality} discusses quality metrics and the impact on generative performance when using the digital neural sampler and sparse network connectivity; Section \ref{sec:truenorth_rbm} shows the developed 3-stage RBM architecture and the generative model on TrueNorth; Section \ref{sec:spike_flow} illustrates the spike processing flow in the TrueNorth RBM; Section \ref{sec:automation} details the design automation procedure for optimal hardware utilization; Section \ref{sec:results} presents the results of the physically-implemented TrueNorth generative RBM; and the last section discusses conclusions and future work.


\section{Markovian analysis of the digital neural logistic sampler}
\label{sec:markov}

The kernel of the MCMC procedure for inference in an RBM is the Gibbs Sampler and involves sampling from a logistic function (Eq. \eqref{eq:logistic}) \cite{geman1984stochastic,smolensky1986information}. More specifically, it involves sampling from a Bernoulli distribution (defining the state of the RBM unit, $x$ in Eq. \eqref{eq:logistic}) parameterized by a logistic function (activation probability). Traditional methods for realizing a logistic sampler in hardware demand a look-up table or functional approximation for the sigmoid \cite{tommiska2003efficient,tisan2009digital,lakshmi2013survey}, which is then compared to the output of a pseudo-random number generator. On the other hand, in spiking neural hardware, such as TrueNorth, the only computational primitives are neurons. Since sigmoidal activation functions are not inherently present in TrueNorth, we therefore have to make use of the deterministic and stochastic neurodynamical properties of the system for efficient realization of the logistic sampler. Below we describe the process of Gibbs sampling using digital spiking neurons in a Markov chain framework, which is useful for better understanding the sampler behavior and serves as a means for producing the generative performance metrics detailed in Section \ref{sec:quality}. The solution neatly combines producing the logistic function and sampling the state of the RBM unit.

In \cite{das2015gibbs} it was shown that a digital integrate-and-fire neuron with a uniformly-sampled threshold combined with a Bernoulli-sampled leak can approximate a logistic spiking probability for the corresponding RBM unit. Here we expand on this by providing a Markov chain analysis of the discrete-time neural sampler. The neural sampling procedure is initialized by setting the neural membrane potential ($V_m$) to a value equivalent to the argument of the logistic function (which is a function of the weights, bias and unit states, as shown in Eq. \eqref{eq:logistic}). Afterward, the system uses three neural variables (two stochastic and one deterministic) to produce an approximate sigmoidal spiking probability. These variables are explained next.

\vspace{5pt}
1) \ \textbf{Stochastic leak.} The stochastic leak is an integer value added to the membrane potential, and is sampled from Bernoulli trials with $p=0.5$. In other words, at every time step (``tick''), either the membrane potential remains the same or it is incremented by the leak value ($L$). This type of leak is inspired from the TrueNorth system, whose neurons can be configured with stochastic non-voltage-dependent leak. For our setup, we chose to use a positive leak, however TrueNorth neurons can take on positive or negative leak values. The TrueNorth system will be explained in detail in Section \ref{sec:truenorth}.

\vspace{5pt}
2) \ \textbf{Stochastic threshold.} The stochastic threshold is an integer sampled from a uniform distribution between $V_{th}$ and $V_{th}+TR$; the term $TR$ stands for ``threshold range''. At every tick, the membrane potential of the neuron is compared with the stochastic threshold and, in case the potential hits (i.e. is equal to or exceeds) the threshold, a spike event will be generated.

\vspace{5pt}
3) \ \textbf{Sampling time window.} The deterministic component of the sampler is the sampling time window ($T_S$), which is the number of ticks during which the neuron is observed. The operation of the sampler during $T_S$ is the following:
	
	\begin{enumerate}[label=\alph*.]
		\item{If during $T_S$ the neuron hits the threshold at least once, a spike event after $T_S$ is produced.}
		
		\item{Even if the neuron hits the threshold more than once during $T_S$, the sampler must still produce a single spike at the output after $T_S$.}
				
		\item{If no threshold events occur during $T_S$, then no spike event is produced at the output of the sampler after $T_S$.}

	\end{enumerate}
	
\vspace{-5pt}
\subsection{Sampling algorithm using digital neurons}
\label{subsec:algorithm}

The algorithm, using TrueNorth-based I\&F neurons with stochastic leak ($L$) and threshold ($V_{th\_rand}$), for realizing the sigmoidal sampling rule (Eq. \eqref{eq:logistic}) to perform MCMC sampling in RBMs is given below.%

\vspace{-3pt}
\begin{algorithm}
	 $V_m = V_{init}$ \\
   $spiked = 0$ \\ 
   \Repeat{$T_S$ steps}{
  $V_m = V_m + $ B(0.5)*$L$ \\
  $ V_{th\_rand} = $ U$(V_{th},V_{th}+TR)$ \\
  \bf{if} $ (V_m \ge V_{th\_rand})$:\ $spiked = 1$}
\end{algorithm}
\vspace{-5pt}

The term B($p$) represents a Bernoulli sample (0 or 1) with probability $p$ and U($a,b$) is an integer sampled from a uniform distribution between $a$ and $b$ (both inclusive). The membrane potential ($V_m$) is initialized to $V_{init}$ and, during the ``repeat'' cycle, if $V_m$ crosses the threshold (equivalent to $V_m \ge V_{th\_rand}$), the $spiked$ variable will be set to 1, after which it will remain in this state until the end of the $T_S$ time steps. Therefore, the state of $spiked$ after $T_S$ ticks will produce a sample (given the initial membrane potential) from an approximate sigmoidal spiking probability distribution. The state of a sampled RBM unit using this algorithm is equal to the state of $spiked$. How to generate the $spiked$ variable using TrueNorth neurons will be explained in Subsection \ref{subsec:quantization}, along with implementation details in Section \ref{sec:truenorth_rbm}. Next, we will analyze the effect of the stochastic neural variables using discrete-time Markov chains.

\subsection{Adaptation of neural variables into Markov chains}

Since we are dealing with a discrete-time digital system, the stochastic neural variables can be modeled as coupling between two discrete-time Markov chains (DTMC): a stochastic leak DTMC and a stochastic threshold DTMC. The sampling time window determines how many steps should be taken in these chains. Each state in a chain is the instantaneous value of the membrane potential. Due to a limited number of bits for data representation in the digital system, saturation levels should be taken into account. For illustrative purposes, in our examples we consider only positive leak values, implying that only the positive saturation level will come into effect, as any data point (i.e. membrane potential) beyond it will be clipped to the saturation value.

The sampler operates by first initializing the stochastic leak DTMC at the state which represents the initial membrane potential value, and then taking alternate steps between the stochastic leak and the stochastic threshold DTMCs. Different initialization values of the stochastic leak DTMC yield different sigmoidal probabilities. Both DTMCs present the same number of states, defined by the membrane potential range. In terms of structure, the DTMCs will always present states representing lower-valued membrane potentials to the left of the chain, and consequently the rightmost state represents membrane potential equal to $V_{sat}$. Next, we discuss the effect of the three neural properties on the DTMCs.

\begin{figure}[h]
	\centering
	\includegraphics[width=0.48\textwidth]{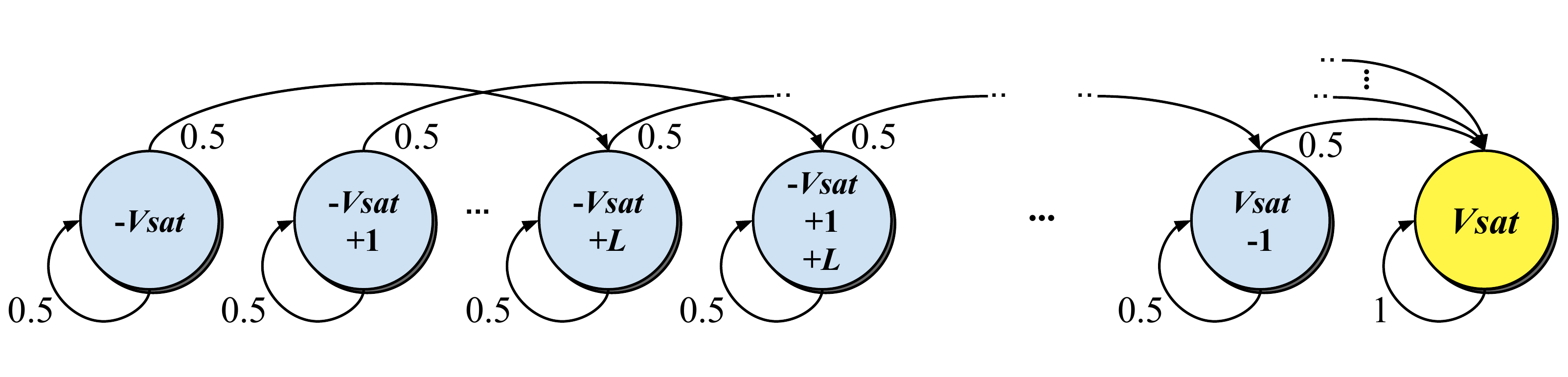}
	\caption{DTMC for stochastic leak in the neural sampler.}
	\label{fig:dtmc_leak}
\end{figure}

\vspace{+5pt}
1) \ \textbf{Stochastic leak DTMC.} Since the stochastic leak chosen for our examples causes only non-negative change in the membrane potential, the only possible transitions, at each stochastic leak tick, from a state are: (1) to itself (in the event of no leak occurrence) or (2) to the right (positive additive leak occurrence). Figure~\ref{fig:dtmc_leak} shows the general case of the DTMC for the stochastic leak ($L$). The number next to each state transition is the transition probability (set to 0.5 for all states). Note how no value of membrane potential can surpass $V_{sat}$, which makes the state representing this specific membrane potential an absorbing state \cite{book_markov1960,clark2008terminating}. Since it is the only absorbing state in the chain, it is called the terminating state in a terminating DTMC. Also note this state will be reached by more than one other state (not considering the self-connection) when $L>1$.

\begin{figure}[h]
	\centering
	\includegraphics[width=0.48\textwidth]{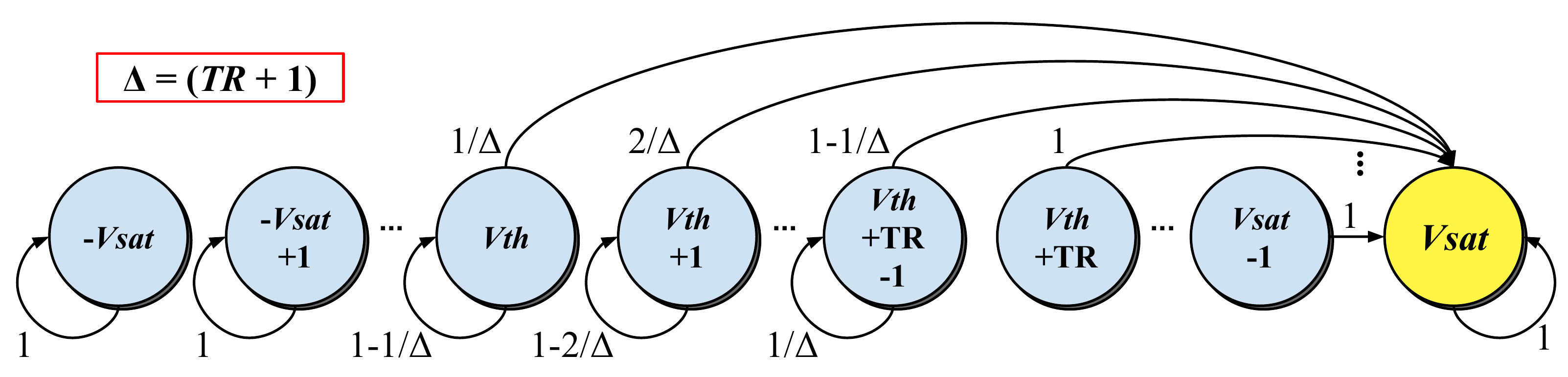}
	\caption{DTMC for stochastic threshold in the neural sampler.}
	\label{fig:dtmc_threshold}
\end{figure}

2) \ \textbf{Stochastic threshold DTMC.} The stochastic threshold is sampled, at each stochastic threshold tick, from a uniform distribution, which produces a linearly increasing transition probability from states inside the range $[V_{th}:V_{th}+TR]$ to the spiking state. Figure~\ref{fig:dtmc_threshold} shows the general case of the DTMC for the stochastic threshold. Note how values outside the range previously described are guaranteed not to hit the threshold when $V < V_{th}$ (realized by the self-connections) and guaranteed to hit the threshold when $V \ge (V_{th}+TR$) (realized by the connections to $V_{sat}$). For simplification, in the figure the symbol $\Delta$ = ($TR$ + 1).

To transform spikes into probabilities, we must produce a single spike event after $T_S$ in case the neuron reached the threshold during $T_S$. This can be obtained by using the $V_{sat}$ state as the terminating state also for the stochastic threshold. A two-fold effect is produced by this terminating state: (1) the two DTMCs become coupled by using a common terminating state; and (2) the sigmoidal firing probability can be extracted directly from the terminating state in the stochastic threshold DTMC after $T_S$, as will be shown below.

\vspace{+5pt}
3) \ \textbf{Sampling time window.} The deterministic component of the sampler, the sampling time window, defines the number of steps taken in the Markov chains and is a two-phase process. The first phase occurs in the leak DTMC, where a new membrane potential value is assigned to the neuron. The second phase is the evaluation of the newly-assigned membrane potential in relation to the noisy threshold. This entire process is considered one step in the coupled DTMCs. In case the system resides in the terminating state, $V_{sat}$, of the coupled DTMCs after $T_S$, an ultimate single spike event will be produced; if the system is in any other state, no spike event will be produced. This results in spike events sampled from the sigmoidal spiking probability, conditioning on the starting state ($V_{init}$) of the procedure.


\subsection{Matrix representation of Markov chains}


A terminating Markov chain presents a single absorbing state, also known as the terminating state; all the other states are transient. The transition probability matrix -- with rows representing origin states and columns representing destination states -- of a terminating Markov chain can be defined in the following manner:

\begin{equation}
\begin{aligned}
		\includegraphics[width=0.1\textwidth]{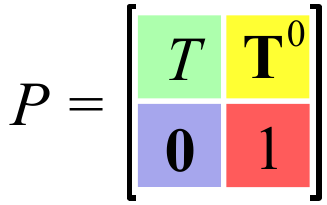}
		\label{eq:P}
\end{aligned}
\end{equation}


\noindent In matrix $P$, the $m \times m$ transient-states transition matrix is represented by $T$, the row-vector \textbf{0} represents the terminating state's non-transient transitions, and ($I_m - T$)\textbf{1} = \textbf{T\textsuperscript{0}}. Therefore, the entire transition matrix $P$ can be characterized by simply knowing $T$.

\vspace{+5pt}
1) \ \textbf{Stochastic leak DTMC.} The stochastic leak is characterized by the additive leak value ($L$). The leak DTMC can be defined by the transition matrix $P_l$ in Eq. \eqref{eq:Pl}. The colors of the matrix components represent the same individual components as in Eq. \eqref{eq:P}.


\begin{equation}
\begin{aligned}
		\includegraphics[width=0.4\textwidth]{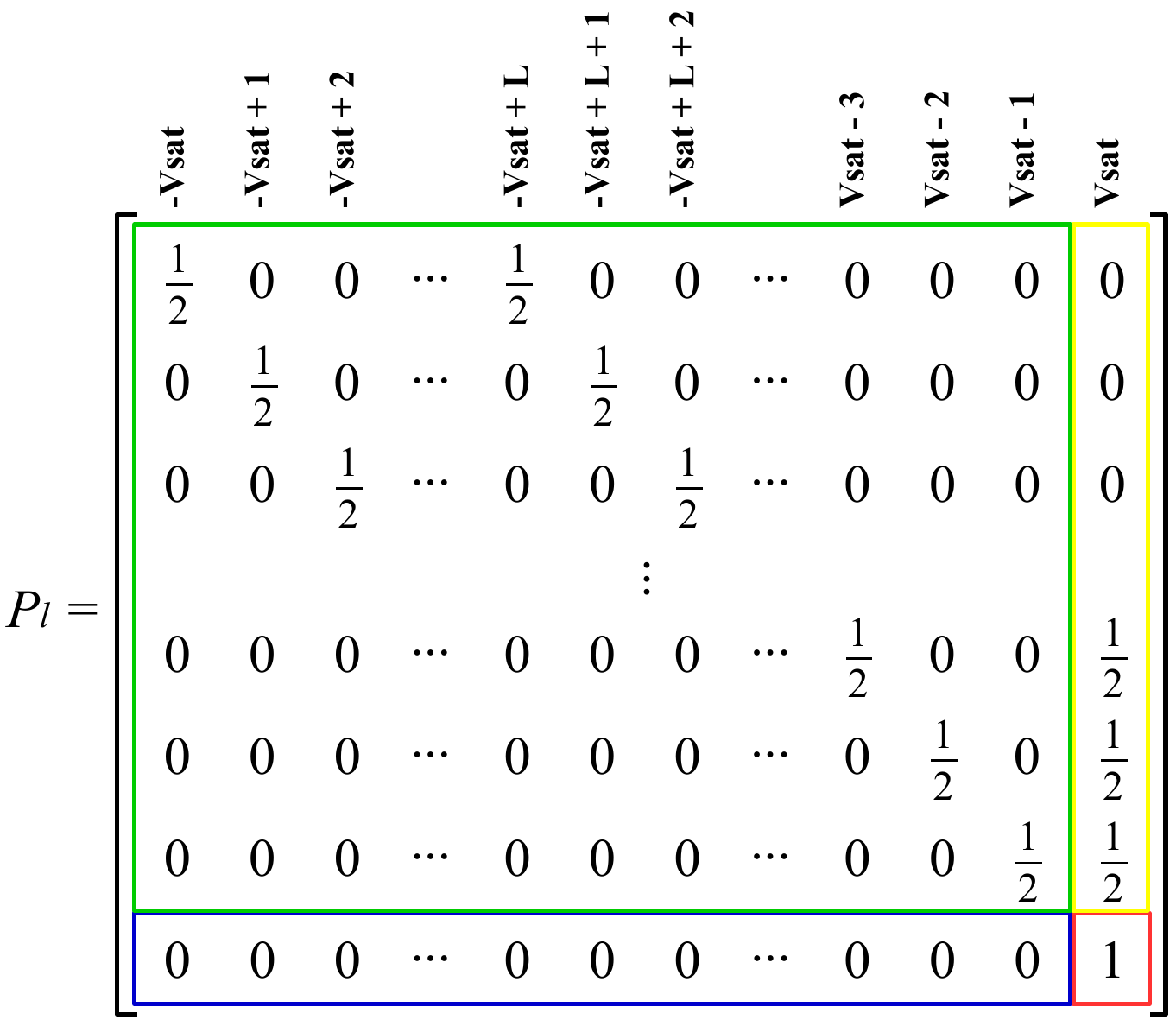}
		\label{eq:Pl}
\end{aligned}
\end{equation}

\vspace{+5pt}
2) \ \textbf{Stochastic threshold DTMC.} The stochastic threshold is characterized by the base threshold value ($V_{th}$) and the threshold range ($TR$). The threshold DTMC can be defined by the transition matrix $P_{th}$ in Eq. \eqref{eq:Pth}. The colors represent the same individual components of $P_{th}$ as in Eq. \eqref{eq:P}.


\begin{equation}
\begin{aligned}
		\includegraphics[width=0.4\textwidth]{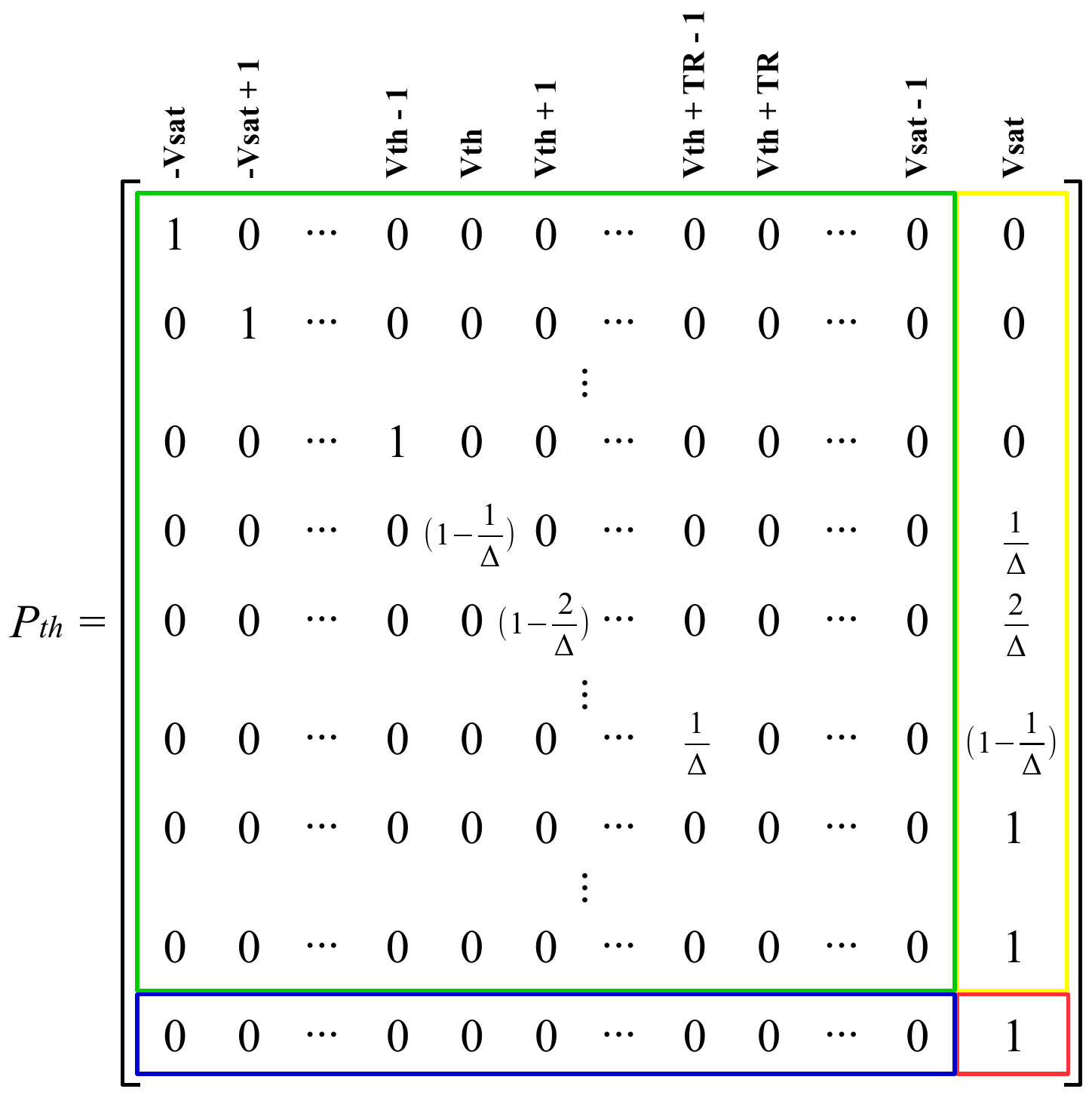}
		\label{eq:Pth}
\end{aligned}		
\end{equation}

\vspace{+5pt}
3) \ \textbf{Spiking probability: coupled DTMCs and sampling time window.} To obtain the sigmoidal spiking probability, the two transition matrices must first be coupled to produce $P_{c} = P_l P_{th}$. The spiking probability can now be obtained by computing $P_{sample}$ = $P_{c}^{T_S}$, which represents $T_S$ steps taken in the coupled DTMC. With this, the last column (terminating state) of the final matrix will contain the spiking probability, $P_{spike}$, of each initial membrane potential (rows in the matrix). Therefore,

\begin{equation}\label{eq:Pspike}
	P_{spike}(s_i) = P_{sample}(s_i,2 V_{sat}+1),%
\end{equation}
where $s_i$ is the origin state corresponding to the initial membrane potential of the neuron prior to sampling.


\subsection{Example}

\begin{figure}[h]
	\centering
	\includegraphics[width=0.49\textwidth]{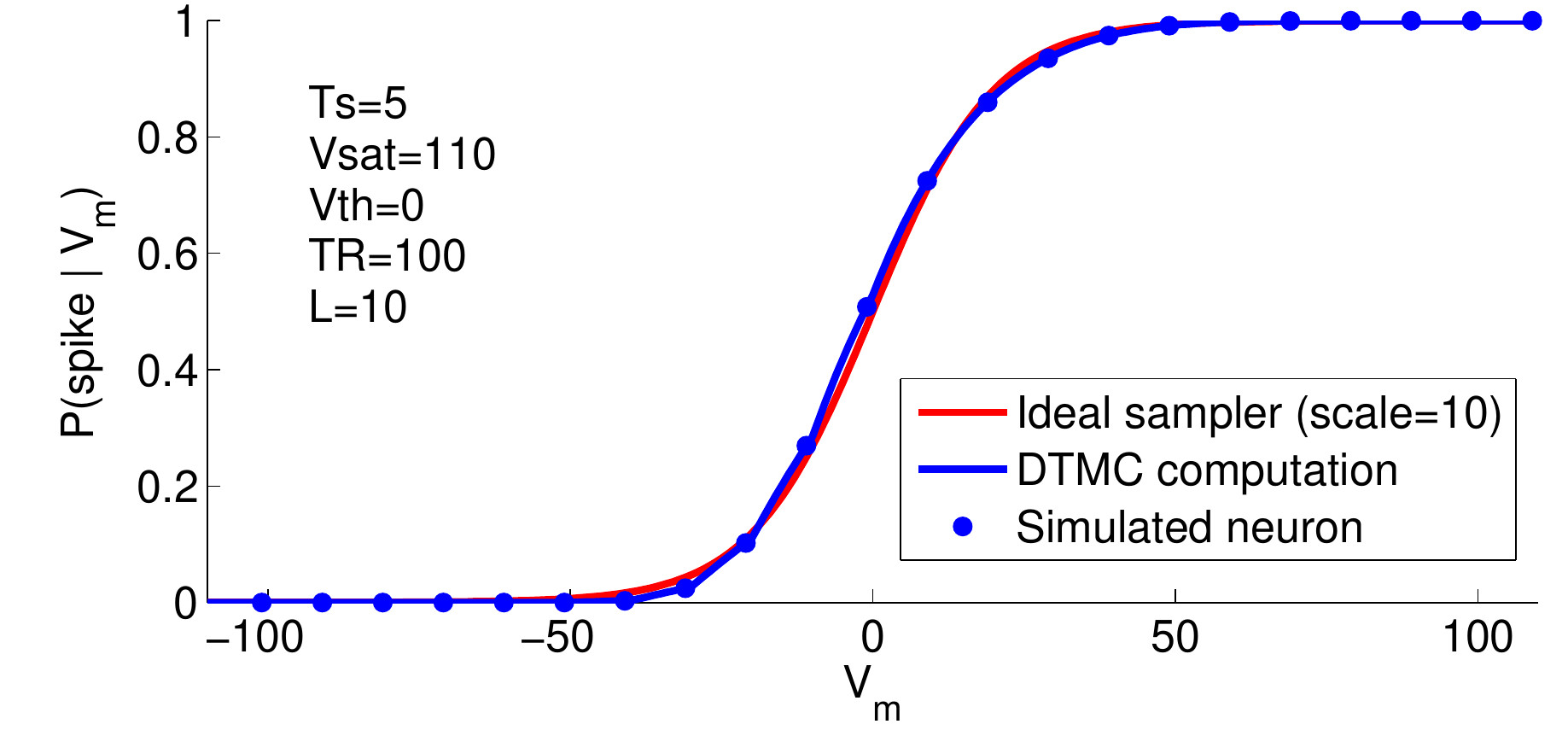}
	\caption{Ideal sampler versus DTMC computation and neural simulation.}
	\label{fig:dtmc_ex2}		
\end{figure}

The example, shown in Fig. \ref{fig:dtmc_ex2}, illustrates the sampler obtained using the previous calculations and compared with actual stochastic neuron simulations. The $x$-axis represents the membrane potential ($V_m$) of the neuron at the start of the sampler operation. As can be seen, the neural sampler obtained via the coupled DTMC computation (blue line) and the stochastic simulation (averaged over $10^4$ samples for each initial $V_m$) of the neuron (blue circles) are overlapping. Besides this, the results from the DTMC computation approximate the ideal sampler scaled by a factor of 10 (red line) with considerable precision.

Since the logistic function in Eq. \eqref{eq:sigmoid} naturally presents a dynamic range between -6 and +6, and because the TrueNorth system deals only with integer-valued membrane potentials, the scaling factor is a means of increasing the resolution of the neural sampler. In other words, by ``streching out'' the function, each integer increment in the initial value of the membrane potential represents a smaller step in the function, resulting in higher resolution. To realize this using the digital neuron model described thus far, the appropriate values of $T_S$, $V_{th}$, $TR$, and $L$ must be chosen. In Subsection III-B, when physically implementing the neural sampler algorithm in the TrueNorth system, the use of the scaling factor is further discussed, and a quantitative analysis of the TrueNorth neural sampler versus the ideal sampler for parameter selection is detailed in Section \ref{sec:quality}.

The noise sources of the stochastic leak and threshold are, respectively, Bernoulli and uniform. By applying these noise sources in a single tick, it is not possible to obtain the precise S-shape of Fig. \ref{fig:dtmc_ex2}; only straight lines could be obtained. Therefore, an explanation for the ``curved'' part of the sigmoid (around $V_m$ equal to -35 and +35) is the non-linear behavior produced by the temporal aspect of the sampler ($T_S$). Throughout multiple ticks, the combination of these ``linear'' noise sources results in a more non-linear curve by creating shorter segments from the straight lines.

\subsection{Considerations}

The problem was analyzed for a positive additive stochastic leak, yet the same would be possible with a negative leak. The main detail is that the last column of $P_{th}$, originally considered the terminating state, would be able to be transitioned out of due to the stochastic leak in the next step of the Markov chain. Also, the terminating state in $P_l$ would be the first column, which does not ``line up'' with $P_{th}$.

On the other hand, if a negative leak is applied, though not sufficient for a chain in the rightmost state to reach a state below the maximum value of the threshold (i.e. below $V_{th} + TR$), then the correct spiking probability can be obtained. In this manner, even if the leak causes a transition to the left in $P_l$, the following iteration of $P_{th}$ will force the system to return to the rightmost state. Interestingly, the coupled activity of the two DTMCs can preserve the original terminating state, even if it is not the terminating state in $P_l$.

Discrete phase-type distributions (DPTDs) \cite{latouche1999introduction} are very similar in nature to the developed neural sigmoid sampler. The main difference is that DPTDs result from a system of one or more inter-related and sequentially occurring geometric distributions, while the neural sampler results from a combination of geometric (leak as Bernoulli trials) and uniform (threshold) distributions.

Lastly, the digital neural sampler is an elegant solution for sampling from a logistic function by not only using bio-inspired neural dynamics but also simultaneously realizing two operations: computing the spiking probability and sampling to obtain the new state of the unit. The DTMC presented can be very useful when simulating the network dynamics: the neuron's transition operator can be extracted by simply accessing the spiking probability curve obtained from the DTMC. This removes the demand of having to simulate every step of the neuron during the sampling time window ($T_S$), which comes in handy during the analysis of the generative performance of the sampler in Subsection \ref{subsec:quality_sampler}.
 

\section{Approaches for Deep Learning\\on TrueNorth}
\label{sec:truenorth}

Neuromorphic substrates present unique challenges for creating spiking versions of machine learning algorithms due to data precision and network connectivity constraints. In this work, a step-by-step methodology for porting RBMs and DBNs onto the IBM TrueNorth system is detailed. The MNIST dataset, consisting of 28x28 pixel grayscale images of handwritten digits 0 through 9, was chosen for the generative inference task. For our experiments, the images were binarized to zero-one values for adaptation to the neuromorphic scenario. The following subsections present the TrueNorth system and outline the approaches and quantitative analysis of the algorithm adaptations necessary for mapping the (offline-trained) networks.


\vspace{-8pt}
\subsection{The TrueNorth digital neurosynaptic processor}
\label{subsec:truenorth}

\begin{figure}[hbtp]
\centering
	\begin{subfigure}{.49\textwidth}
		\centering
		\includegraphics[width=0.90\textwidth]{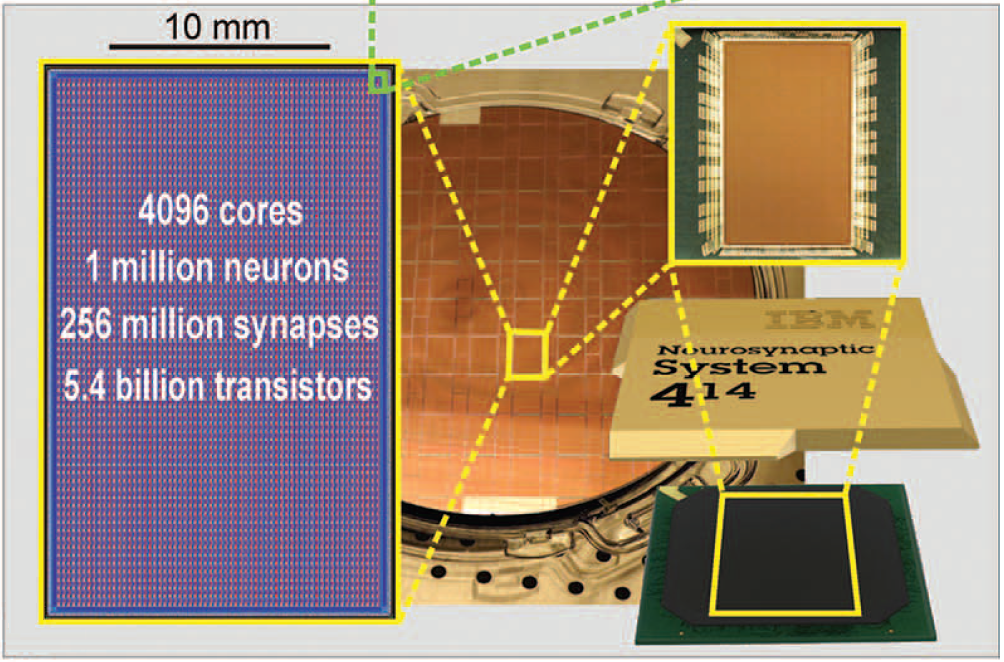}
		\caption{}
		\label{fig:tn_a}
	\end{subfigure}\\
	\vspace{+5pt}
	\begin{subfigure}{.24\textwidth}
		\centering
		\includegraphics[height=3.4cm]{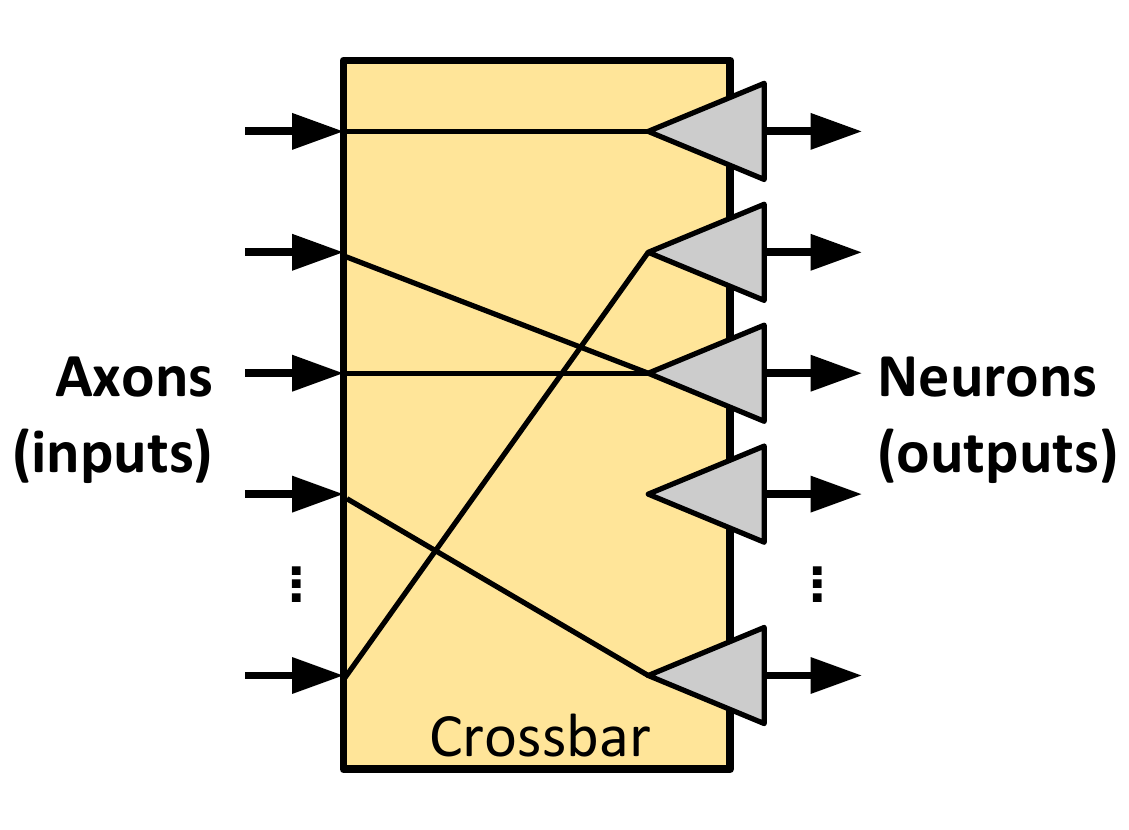}
		\caption{}
		\label{fig:tn_b}		
	\end{subfigure}%
	\begin{subfigure}{.24\textwidth}
		\centering
		\includegraphics[height=3.4cm]{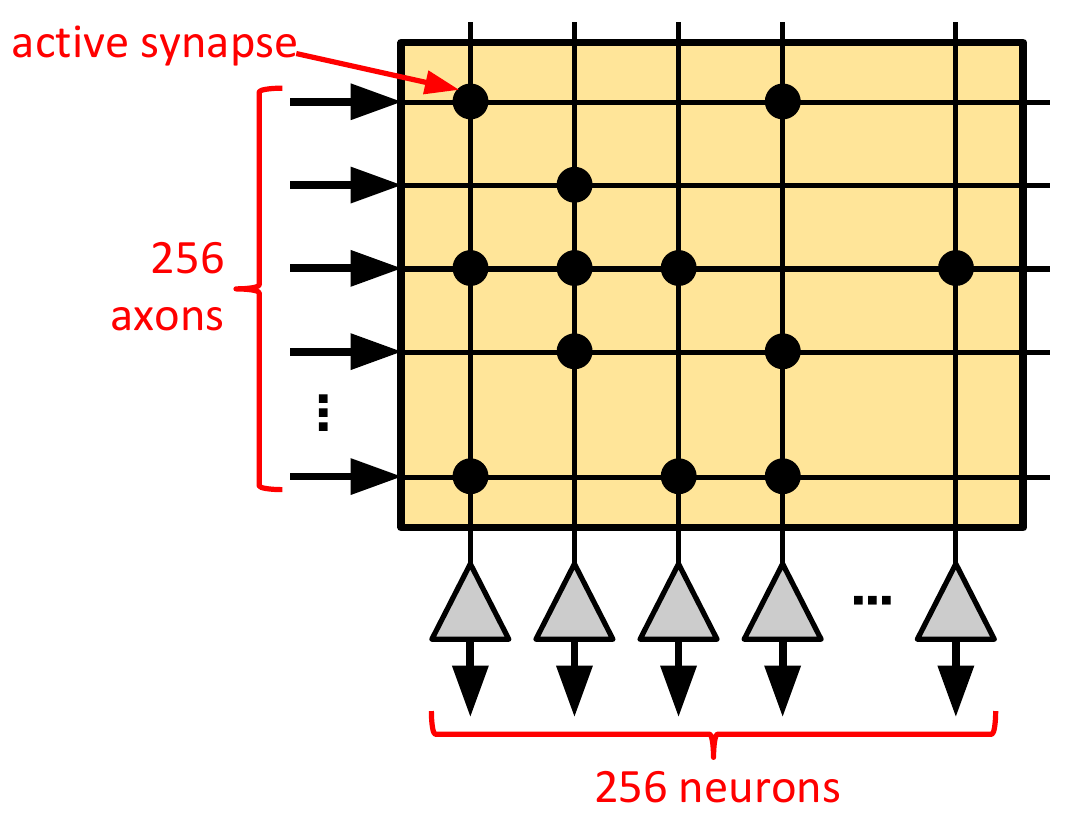}
		\caption{}
		\label{fig:tn_c}		
	\end{subfigure}%
	\caption{The TrueNorth neurosynaptic processor: (a) chip layout, wafer, and chip package; (b) high-level view of the 256 axons (inputs) and 256 neurons (outputs); and (c) internal view of the fully-configurable binary crossbar \cite{merolla2014million}.}
\end{figure}

IBM's TrueNorth is a very low-power, brain-inspired digital neurosynaptic processor \cite{merolla2014million}, with 4096 cores, totaling 1 million programmable spiking neurons and 256 million configurable synapses (Fig. \ref{fig:tn_a}). The core is the basic building block of the system, each composed of 256 axons (inputs) and 256 neurons (outputs) (Fig. \ref{fig:tn_b}), connected via a 256 x 256 crossbar of configurable synapses (Fig. \ref{fig:tn_c}). Each neuron can target its generated spikes to any axon on the chip, limited to one axon per neuron, and presents over 20 individually programmable features, including threshold, leak, reset, and stochastic properties. From the user's point-of-view, neurons operate in 1 ms time steps, during which asynchronous spike event transmission and processing occurs between and inside the cores. Therefore, during each 1 ms interval, spikes are delivered to and processed in their destination cores, after which a global clock aligns the generation of the next set of spikes.


The digital integrate-and-fire (I\&F) TrueNorth neurons present stochastic and deterministic leak and threshold properties. A simplified representation of the dynamical behavior of the membrane potential $V_{j}(t)$ for neuron $j$ at time $t$ is defined by the following set of (sequentially processed per neuron) equations \cite{Cassidy13cognitivecomputing}:
\begin{subnumcases}{}
V_{j}(t) = V_{j} (t-1) +\Sigma_{i=0}^{255} \ A_{i}(t) \ w_{i,j} \ s_{j}^{G_i} \label{eq:tn1}\\
V_{j}(t) = V_{j}(t) + (1-c_j)\lambda_{j}+ c_j F(\lambda_j) \textrm{sgn}(\lambda_j) \label{eq:tn2} \\
\textrm{if} \ (V_{j}(t) \ge \alpha_{j} + \eta(M_j)), \ \textrm{Spike and set } V_{j}(t) = R_{j} \ \ \ \textrm{   } \ \ \ \label{eq:tn3}
\end{subnumcases}


\vspace{+3pt}
The first line (Eq. \eqref{eq:tn1}) represents the synaptic integration of all active axons impinging on neuron $j$ at time $t$. The term $A_i(t)$ is the binary-valued input spike arriving from the $i^{th}$ axon at time $t$; $w_{i,j}$ is the binary-valued synaptic connection between axon $i$ and neuron $j$; and $s_{j}^{G_i}$ is the synaptic weight between axon $i$ and neuron $j$. This last term is particularly interesting as each neuron presents four 9-bit signed integer configurable weights. Therefore, an axon can be configured to be one of four types, and this defines which of the four possible weight values -- individually in each neuron it is connected to -- will be integrated if the axon is active.

The second line (Eq. \eqref{eq:tn2}) represents the leak integration, where $\lambda_{j}$ is a 9-bit signed integer. Depending on the value of $c_j$, the leak can be deterministic ($c_j=0$) or stochastic ($c_j=1$). When $c_j=0$, the value of $\lambda_j$ is integrated in the membrane potential. On the other hand, when $c_j=1$, the stochastic function $F(\lambda_j) = |\lambda_j| \ge \rho$ defines if a leak of value sgn($\lambda_j$) is  integrated; the value of $\rho$ is a sampled uniformly distributed 8-bit integer. In this manner, a stochastic leak can only take on values of +1 or -1. However, the value of $L$ in the digital neural sampler (refer to the algorithm in Subsection \ref{subsec:algorithm}) can take on much larger values. How to implement stochastic leaks greater than 1 on TrueNorth will be explained in Section \ref{subsec:quantization}.

The last line (Eq. \eqref{eq:tn3}) compares the integrated membrane potential with the threshold, which has a base value of $\alpha_j$ and a uniformly sampled value of $\eta(M_j)$ ranging from 0 to $2^{M}-1$. Therefore, if $V_j(t)$ is equal to or surpasses the threshold, the neuron spikes and its membrane potential is reset to $R_j$. Using the TrueNorth system as a basis for digital neural processing, the next section shows how an approximation to the Gibbs Sampler can be obtained using these neural properties.


\vspace{-8pt}
\subsection{Gibbs sampling with TrueNorth neurons}
\label{subsec:quantization}

Neural sampling can be realized on the TrueNorth system by means of the algorithm described in Subsection \ref{subsec:algorithm}. The first step of the algorithm is to set the initial membrane potential of the neuron ($V_{init}$ in the algorithm) to the equivalent value of the argument of the logistic function. This is realized in TrueNorth by appropriately activating the axons of neuron $j$ at time $t=1$ to produce the desired membrane potential ($V_j(1)$ = $V_{init}$ in Eq. \eqref{eq:tn1}). The neuron is then free to run (no axon activity) during a sampling time window, $T_S$, defined in number of 1 ms time steps (``ticks''), during which a stochastic additive leak is applied and the updated membrane potential is evaluated at every tick. If the neuron's membrane potential is greater than or equal to the stochastic threshold (i.e. the neuron spikes) at least once during $T_S$, the binary state of the equivalent RBM unit is set to 1 (i.e. the RBM unit spikes).

For adapting the algorithm in Subsection \ref{subsec:algorithm} to TrueNorth, the stochastic threshold can be directly modeled by setting the appropriate values of $\alpha_r$ and $M_r$ for TrueNorth neuron $r$. The stochastic leak, on the other hand, cannot be directly mapped for absolute leak values greater than 1. An alternative to this is to use an additional neuron $l$ to act as the stochastic leak for neuron $r$. For this, the parameters of neuron $l$ are set to $c_l=1$, $\lambda_l=+128$, $\alpha_l=1$, $M_l=0$, and $R_l=0$. Therefore, neuron $n_l$ naturally spikes with probability $p=\lambda_l / 255 \approx 0.5$, because it leaks sgn($\lambda_l$) = +1 with this same probability and the threshold is set to $\alpha_l=1$. After spiking, it is reset to $V_l=0$ and will present the same behavior in the next tick. If we then connect the output of neuron $l$ to an input axon (of type $i$) of neuron $r$ and set the memory position $s_r^{G_i}$ equal to the leak value $L$, we will obtain the desired spiking behavior.

In the algorithm, the state of the RBM unit is equivalent to that of the $spiked$ variable. However, in the TrueNorth implementation multiple spikes may be produced by sampling neuron $r$ during $T_S$. A solution for this is to create a so-called ``refractory effect'' using an additional neuron $k$. What this additional neuron essentially does is count how many spikes are received from neuron $r$. For this, neuron $k$ is configured with threshold $\alpha_k=1$ and its membrane potential set at the start of the sampling phase (i.e. at the same moment neuron $r$ is set to $V_{init}$) to, for example, $-T_S$. By incrementing the membrane potential by 1 for every received spike from neuron $r$, the membrane potential of neuron $k$ will only be larger than $-T_S$ if at least one spike was received. Therefore, after $T_S$ has expired, we inject an axonal event of $+T_S$ into neuron $k$, which, due to the unit-valued threshold, will cause it to spike if at least one spike was produced by neuron $r$ during the sampling phase.

In sum, the outset of a single RBM unit can be analyzed as capturing the dynamical behavior of two coupled DTMCs run for a time interval of $T_S$, which are used to set a threshold flag ($spiked$). Two TrueNorth neurons ($r$ and $l$) are used to form the DTMCs, along with a third neuron ($k$) used for verifying if the threshold flag has been triggered, after $T_S$ has expired (thus, the ``refractory effect''). The combination of all of this comprises an RBM unit.

\begin{figure}[h]
	\centering
	\includegraphics[width=0.49\textwidth]{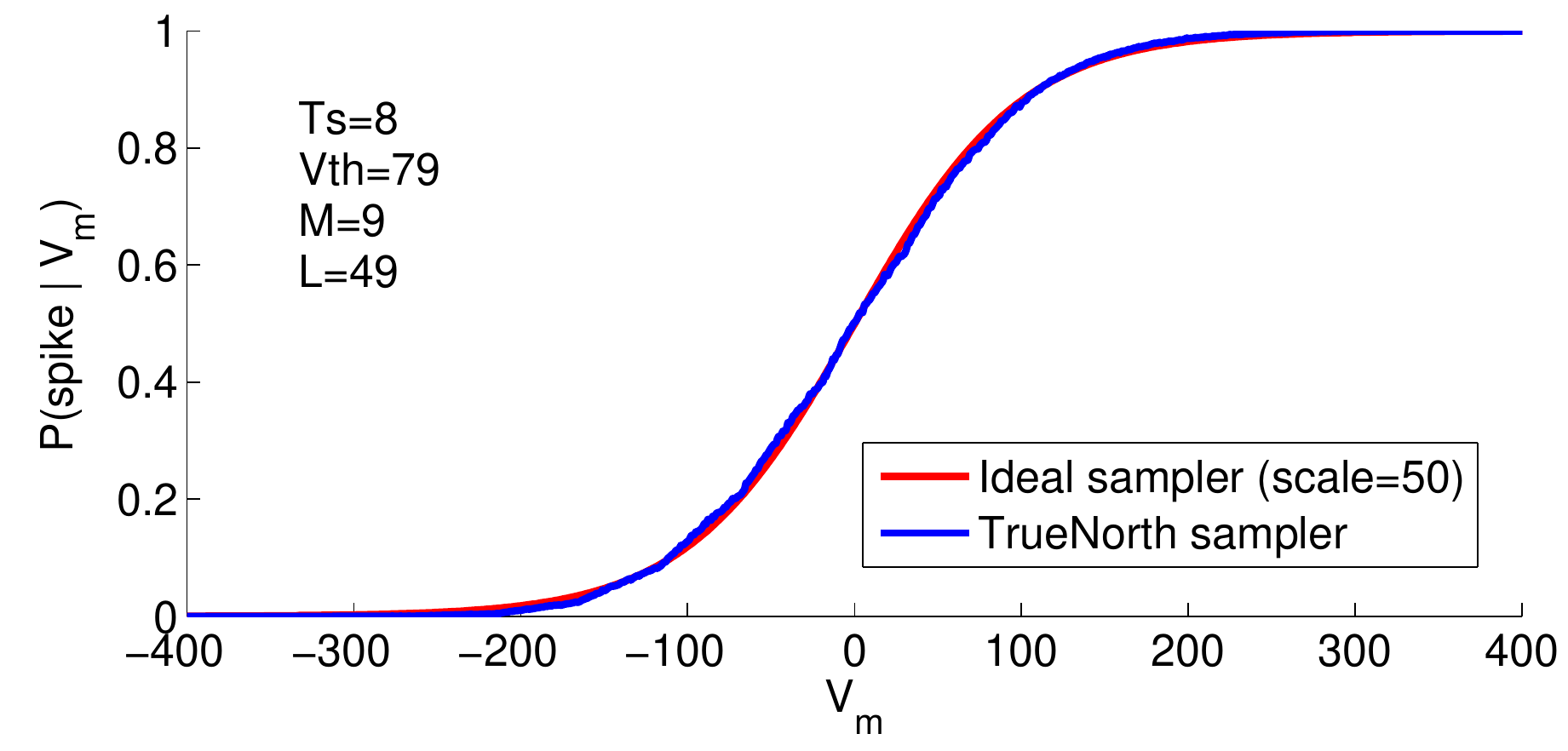}
	\caption{Logistic sampler using TrueNorth neurons.}
	\label{fig:sigmoid}
\end{figure}

As a final note, it was shown that the argument of the logistic function ($x$ in Eq. \eqref{eq:sigmoid}) is modeled as the membrane potential of the neuron. Since TrueNorth neural membrane potential takes on only signed integer values, and the logistic function has a dynamic range between approximately -6 and +6, it is necessary to apply a multiplicative scaling factor, $s$, to the RBM weights and biases to increase the dynamic range of the neural logistic sampler realization. As a result of this scaling, the neural sampler must be realized with appropriate values of $T_S$, $V_{th}$, $M$, and $L$ to enable the RBM to sample with high precision from the logistic probability distribution. The ideal sampler (with $s=50$) is compared with the TrueNorth realization ($T_S=8$, stochastic threshold ranging from 79 to 590, and stochastic leak of 49) in Fig.~\ref{fig:sigmoid}.

\vspace{-10pt}


\subsection{Sparse connectivity}
\label{subsec:sparsity}
\begin{figure}[hbtp]
\centering
	\begin{subfigure}{0.25\textwidth}
		\centering
		\includegraphics[width=0.82\textwidth]{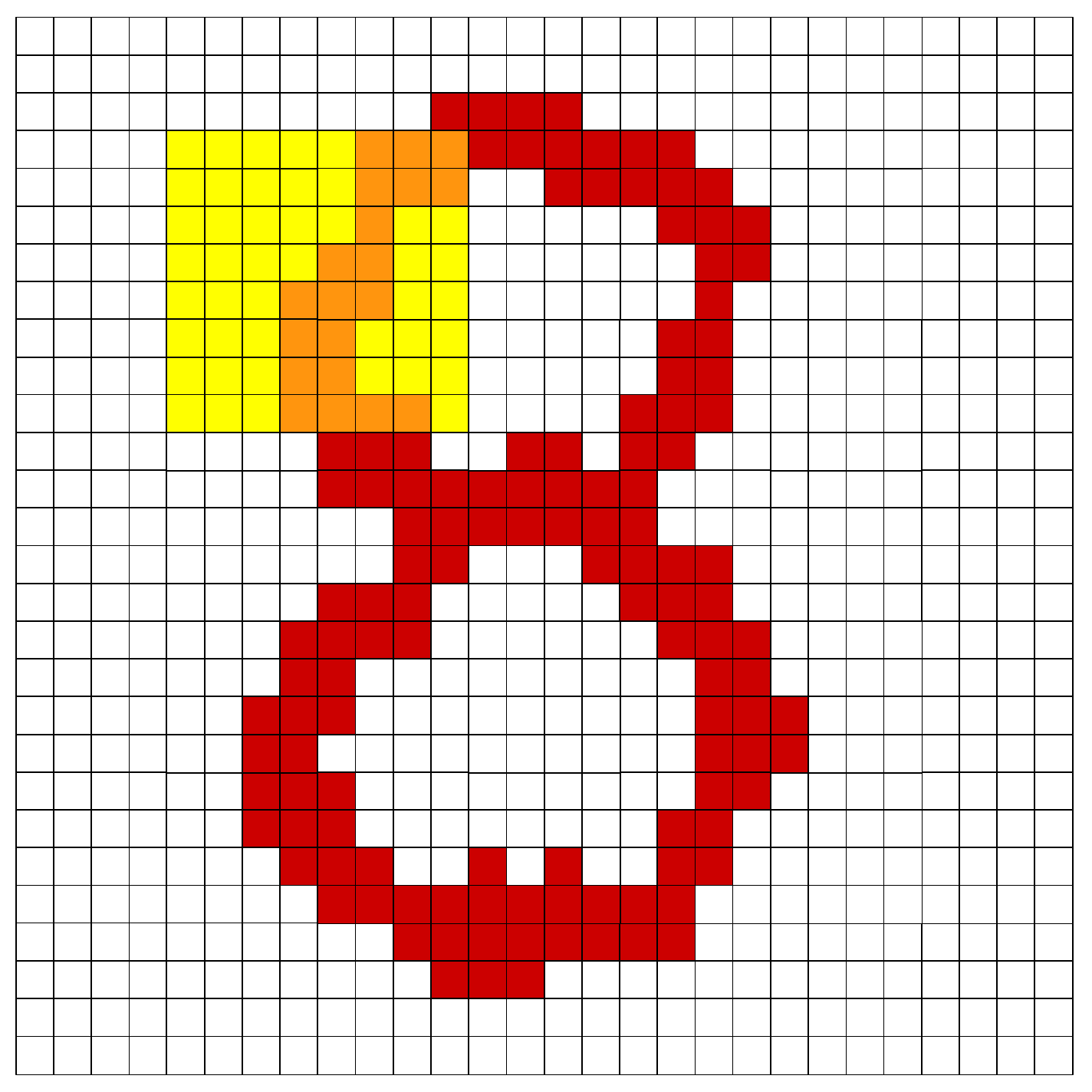}
		\caption{}
		\label{fig:patching_a}
	\end{subfigure}%
	\begin{subfigure}{0.25\textwidth}
		\centering
		\includegraphics[width=0.9\textwidth]{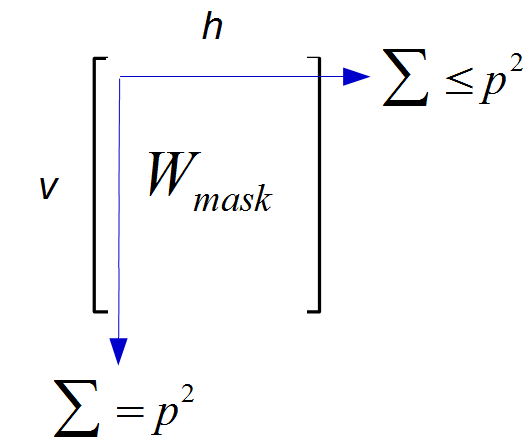}
		\caption{}
		\label{fig:patching_b}
	\end{subfigure}\\
	\caption{Sparsity structure in RBM. (a) Illustration of an 8x8 pixel patch (in yellow). (b) Sparsity can be seen as applying a mask over the network's weight matrix during offline training.}
	\vspace{-4pt}
\end{figure}

The all-to-all connectivity between layers in the RBM algorithm implementation has to be adapted to the available connectivity in hardware. The 256-input cores in TrueNorth present a constraint for the 784-pixel images used in the hand-written digits pattern completion application described in this paper, where each hidden unit, in a standard RBM implementation, is connected to all 784 visible units. A viable solution is to use a patching scheme over the original image \cite{tang2010deep}, thus reducing the area of the image ``observed'' by each hidden unit. Reciprocally, since the generative RBM presents feedback from the hidden layer to the visible layer, the quantity of hidden units should also be selected in a way as to reduce the number of units ``observed'' by the visible units. Fig.~\ref{fig:patching_a} shows how a patch (yellow) is formed by an 8x8 pixel window over a binarized MNIST image.

In \cite{tang2010deep}, square patches of size $p \times p$ were randomly placed over the input image, with all the visible units belonging to a patch connected to a single hidden unit. Though this resulted in reduced network connectivity, for a physical implementation with fan-in constraints a systematic patching scheme is necessary to produce a well established maximum number of units observed in each layer. The systematic patching is particularly important for the generative model due to the feedback from hidden to visible units during inference (details of generative RBM operation are given in Section \ref{sec:truenorth_rbm}). If, for example, patching were performed randomly, a visible unit could possibly be captured by more than 256 patches, making this fan-in unfeasible on TrueNorth. Therefore, we applied an overlapping, yet deterministic, patching scheme developed for the generative RBM realization. The method uses patches with $p^2$ pixels which are formed by ``sliding'' a square window over the $N^2$-pixel image and forming a new patch at every new position. The total number of overlapping patches produced using this method is defined by:

\begin{equation} \label{eq:patch} \textrm{patches = hidden units} = (N-p+1)^2. \end{equation}

A patching scheme implemented in this manner can be interpreted as applying a mask, $W_{mask}$, over the RBM's weight matrix, where 0's and 1's in the mask represent, respectively, no connection and presence of connection between visible and hidden units. The mask is applied during the offline RBM training and the resulting sparse weight matrix is then used for mapping the RBM onto TrueNorth.  In Fig.~\ref{fig:patching_b}, the sum of column values in each row of $W_{mask}$ represents the number of hidden units observed by each visible unit, and the sum of row values in each column represents the number of visible units observed by each hidden unit. With this systematic patching, the bounds of the sums (both in rows and columns) are well-defined.


\section{Quality metrics of digital neural sampler and sparse network}
\label{sec:quality}
The following subsections present quantitative analyses of the impact of the adaptations demanded during the mapping of the original RBM algorithm onto TrueNorth. First, the impact of quantization due to digital hardware data representation is verified. Second, the effect of approximate logistic sampling using the digital neural sampler is analyzed. Lastly, due to non-viability of all-to-all connections between neurons in TrueNorth, we analyze the impact of sparsity in network connectivity. For the neuromorphic adaptations, the generative performances are verified using the Kullback-Leibler (KL) divergence and Annealed Importance Sampling (AIS), which are briefly explained next.\\

\textbf{Kullback-Leibler divergence.} KL divergence is a measure of the difference between probability distributions. The probability distribution of an RBM is defined by Eq. \eqref{eq:energy}, with the denominator of this equation (i.e. the partition function) demanding a countable normalizing sum of all state probabilities for computation. Therefore, since we want to compare the performance of the samplers versus exact probability distributions (computed by Eq. \eqref{eq:energy}), only small networks with tractable partition functions can be analyzed. KL divergence is particularly important for our analysis of the digital neural sampler and, though we cannot directly extrapolate values of this measure to larger networks, the results aid in identifying expected performance for each sampler. KL divergence is defined by the following equation \cite{cover2012elements}:

\begin{equation} \label{eq:kl_div} D_{KL}(P||Q) = \sum_i P(i) \log \frac {P(i)}{Q(i)}, \end{equation}

\noindent where $P$ and $Q$ are two probability distributions, and $D_{KL}$ is always non-negative. The state of the system is defined by $i$. For our experiments, $P$ was defined as the distribution obtained in the experiment and $Q$ as the true distribution.\\

\textbf{Annealed Importance Sampling.} AIS is a metric used to estimate the log-probability of a generative model \cite{neal2001annealed,salakhutdinov2008quantitative}, where larger values indicate higher likelihood that the model generated the data. For high-dimensional models, such as RBMs, where calculation of the partition function is intractable, the AIS algorithm is very useful as it performs a stochastic estimation of the partition function to compute the log probability of the model with respect to the data. Therefore, the AIS algorithm will be used for validating the generative performance of the sparsely connected network by verifying the patch size which produces the largest AIS value.


\subsection{Quality of data quantization}

\begin{figure}[hbtp]
\centering
	\includegraphics[width=9cm]{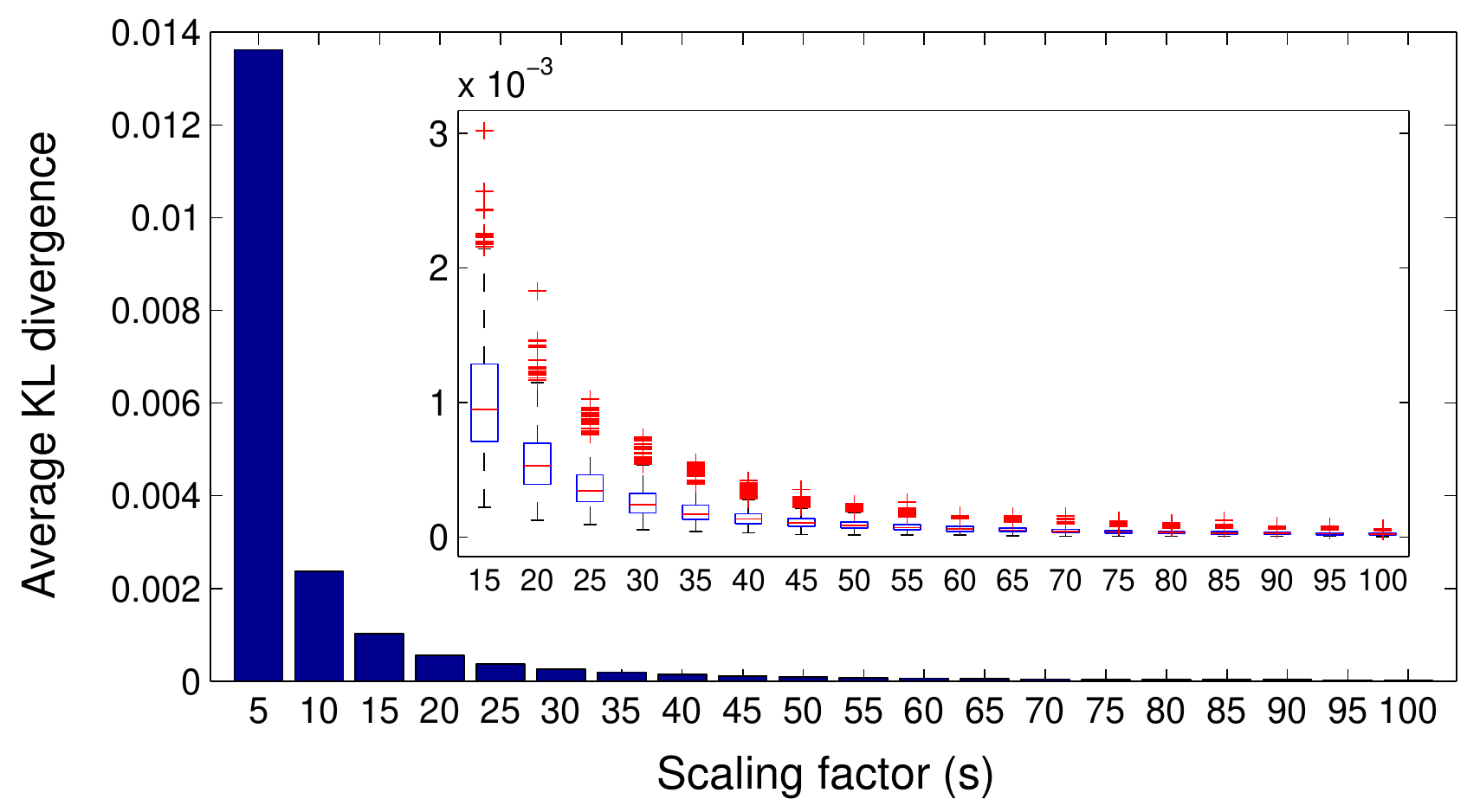}
	\caption{Generative performance versus quantization.}
	\label{fig:kl_div_scaling}
\end{figure}

The effect of data quantization can be verified by comparing the quantized samplers to the ideal. The weights and biases can be quantized by realizing the following: multiply their values by a scaling factor ($s$), then round the result to the nearest integer, and finally divide the second result by $s$. The KL divergence of the network with quantized versus exact (high precision) weights was computed over 1000 experiments, each consisting of randomly generated weights and biases for a network with 5 visible and 5 hidden units. For these, based on experimental results of weights and biases from previously trained RBMs, the values were sampled from the following normal distributions: weights $\sim N(-0.05, 1.6e-3)$, visible biases $\sim N(-0.3, 1)$, and hidden biases $\sim N(0.5, 2.25)$. The KL divergence results of quantized versus non-quantized data are shown in Fig. \ref{fig:kl_div_scaling}, including a box plot of KL divergence for $s = 15-100$. A saturation point can be seen around $s = 50$. It is important to note that very large values of $s$ are beneficial for the algorithm, however they can be costly in terms of hardware resources (cores, in the case of TrueNorth), since more neurons and longer accumulation times will be required for mapping larger values of weights and biases (explained in Section \ref{sec:truenorth_rbm}).

\vspace{-10pt}

\subsection{Quality of the digital neural sampler}
\label{subsec:quality_sampler}

The scaling factor impacts the resource usage of the system and it also impacts the latency -- by increasing the accumulation time (see Section \ref{sec:spike_flow}). The other parameter which also affects latency is the sampling time window ($T_S$). Thus, when neural samplers present the same generative performance, the selected configuration will naturally be the one presenting the lowest $T_S$. Additionally, when selecting the neuron parameters (the ``sampler configuration''), one important aspect of TrueNorth neurons that should be taken into account is the membrane potential range. From Eq. \eqref{eq:tn3}, we can observe that the upper bound (i.e. positive saturation) value of membrane potential is defined by the sum $\alpha_j + \eta(M_j)$. Therefore, for our analysis, we have chosen configurations with this sum close to or surpassing the upper bound of the dynamic range of the sigmoid function ($\approx$ 6$\times$ scaling factor) while still presenting adequate sigmoid fitting.


To begin the neuron parameter selection, we first fix the scaling factor, then the quality of the digital neural sampler can be verified by sweeping over values of sampling time window ($T_S$) and neuron parameters ($V_{th}$, $M$, and $L$) to, ideally, overlap with the sigmoid. The best fit was found for each $T_S$ value by performing a parameter search to reduce the mean squared error (MSE) between the ideal (scaled) logistic function and the curve produced by the neural sampler. For our experiments, five configurations were chosen, with the TrueNorth neuron parameters of the configurations (G1-G5) shown in Table \ref{tb:configs}. The MSE of each versus the ideal logistic function is presented in the rightmost column.

\begin{table}[h!]
\normalsize
  \begin{center}
  \begin{tabular}{| c || c | c c c c | c |}
    \hline
    Config. & Scaling factor & $T_S$ & $Vth$ & $M$ & $L$ & MSE\\
    \hline
    G1 & 50 & 1 & 0 & 7 & 125 & 0.4878 \\
    G2 & 50 & 2 & 0 & 8 & 100 & 0.1311 \\
    G3 & 50 & 4 & 66 & 8 & 77 & 0.0741 \\
    G4 & 50 & 8 & 79 & 9 & 49 & 0.0412 \\
    G5 & 50 & 16 & 186 & 9 & 36 & 0.0415 \\        
    \hline  
  \end{tabular}
  \end{center}
  \caption{Neuron configurations for neural sampler analysis.}
  \label{tb:configs}
\end{table}

\begin{figure}[hbtp]
\centering
	\includegraphics[width=0.49\textwidth]{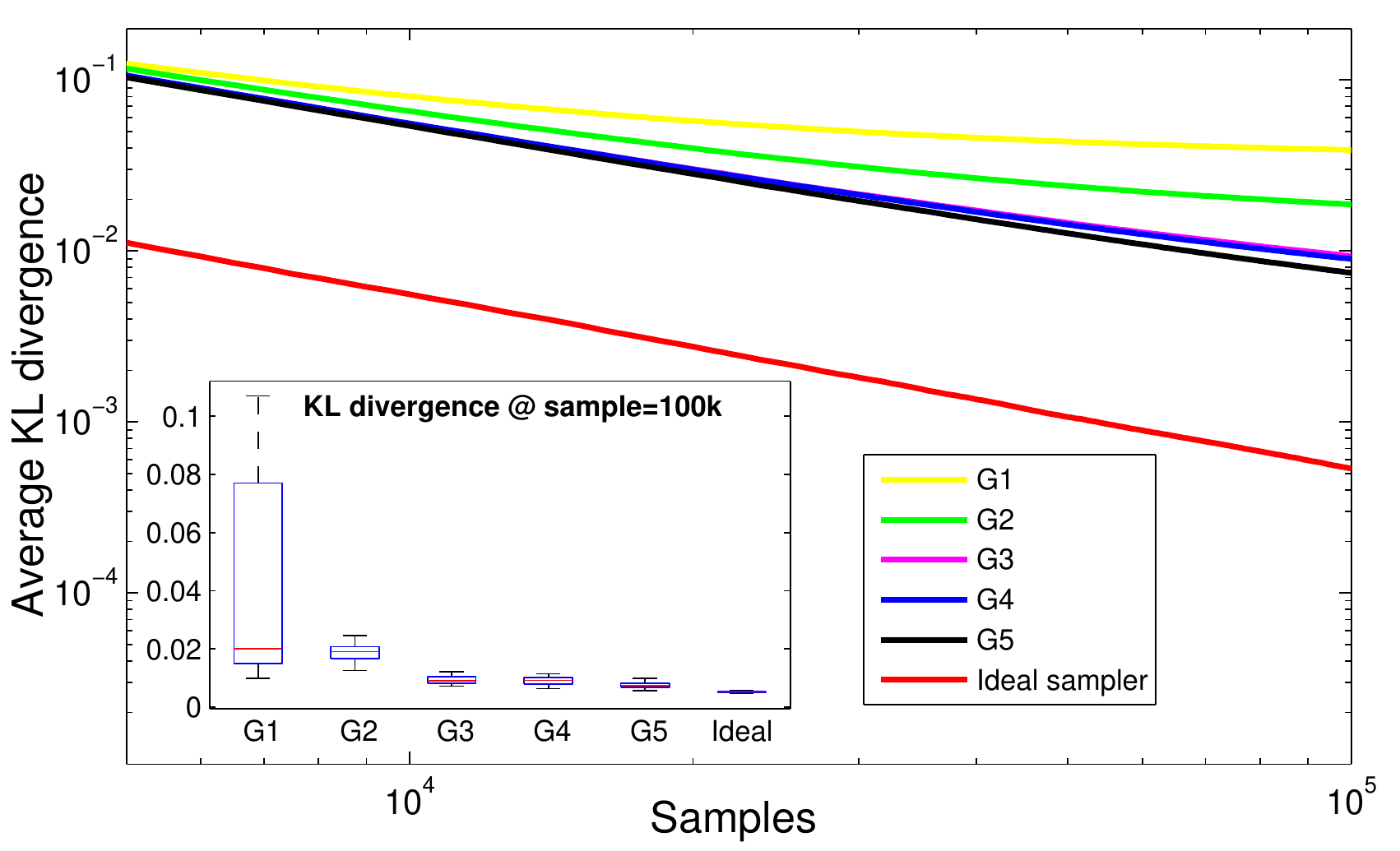}
	\caption{Generative performance of neural Gibbs samplers.}
	\label{fig:kl_div_a}
\end{figure}

The generative model performance for these configurations was determined by means of average KL divergence of the model, and also the ideal logistic sampler (using Eq. \eqref{eq:logistic}), versus the true distribution (computed by Eq. \eqref{eq:energy}), over 10 randomly sampled networks (5 visible and 5 hidden units), with 15 experiments run for each network, and each experiment consisting of $10^5$ samples. Fig.~\ref{fig:kl_div_a} shows the average KL divergence results of the different parameter configurations. The smaller plot in this figure is a boxplot of the 150 (10 networks $\times$ 15 experiments per network) KL divergence values at sample $10^5$. Naturally, the configurations with lower MSE also presented lower KL divergence, with G3 and G4 practically overlapping.

\vspace{-5pt}

As was mentioned in the end of Section \ref{sec:markov}, the DTMC computations of the neural sampler can be very useful when simulating the network dynamics. Instead of having to simulate every step of the neuron during the sampling time window ($T_S$), we can simply use the spiking probability curve obtained from the DTMC as the neuron's transition operator. In other words, the probability of spiking after $T_S$ can be extracted from the curve and this value is then compared to a uniformly-sampled number between 0 and 1. Though this does not affect in any sense the operation of the neural sampler algorithm (and cannot be used in practice), it speeds up simulations considerably.

\begin{figure}[hbtp]
\centering
	\includegraphics[width=0.49\textwidth]{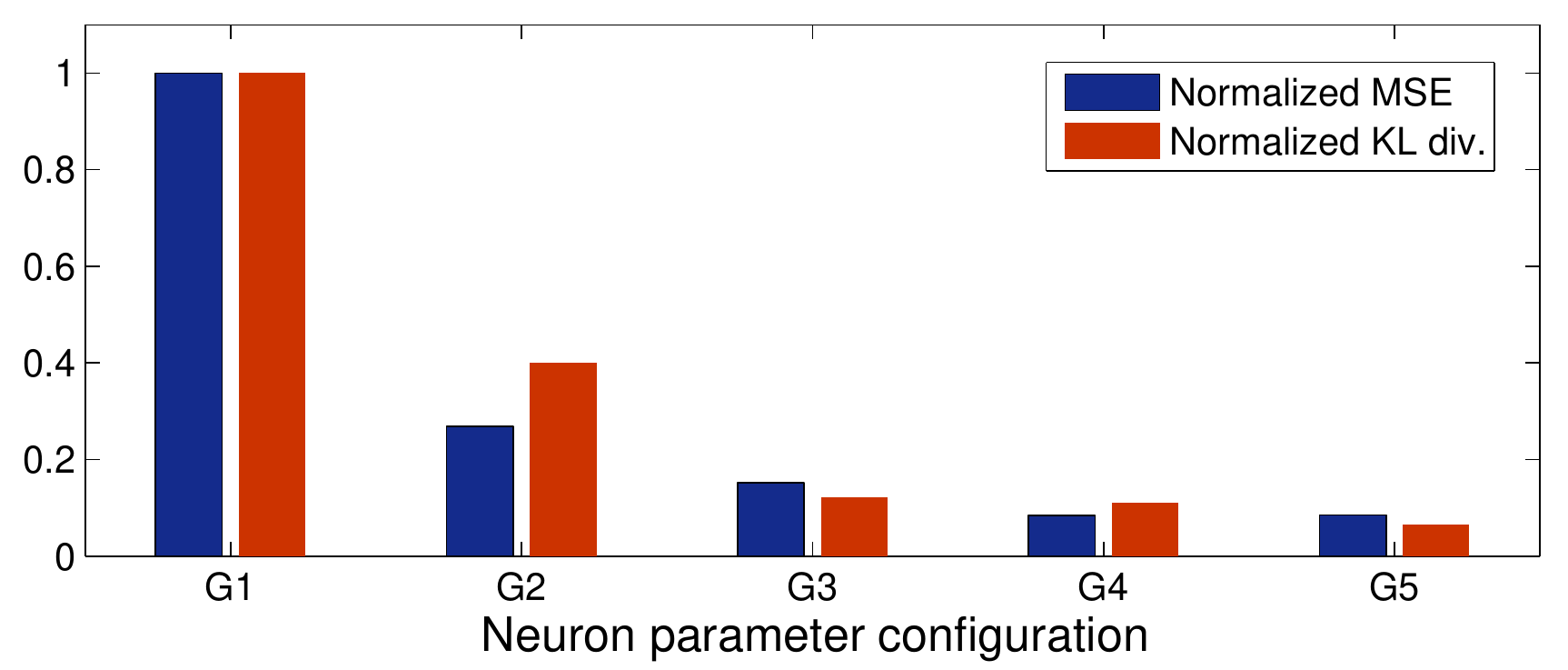}
	\caption{MSE and KL divergence of neural Gibbs samplers.}
	\label{fig:kl_div_b}
\end{figure}

A comparison of the normalized (i.e. all values divided by the worst case = model G1) MSE and KL divergence (at sample $10^5$) is shown in Fig. \ref{fig:kl_div_b}. Though the results for both measures were not identical -- for example, the MSE for G4 and G5 were basically identical, yet the KL divergence for G5 showed a slight improvement --, the figure clearly shows similar trends for both measures. Thus, these results indicate that using the DTMC analysis of the sampler combined with the MSE measure can be a powerful tool for quick access to estimating the generative performance of a sampler. For the generative RBM implementation on TrueNorth, configuration G5 was chosen due to slightly better KL divergence results.

\vspace{-10pt}


\subsection{Quality of the sparse network}

Since sparsity is difficult to evaluate in small networks, the generative qualities of the sparse RBM were verified by means of the AIS measure of a network with 784 visible and 500 hidden units, pre-trained using the MNIST dataset. In a related vein, reference \cite{tang2010deep} shows how a sparsely connected RBM can produce a more noise-tolerant model for classification. For our application, sparsity is actually necessary for reducing the fan-in of each neuron, which, in TrueNorth, is limited by the 256-input cores. The patching scheme proposed for the generative RBM, described in Subsection \ref{subsec:sparsity}, takes into account the feedback from hidden to visible units. With this method, as illustrated in Fig. \ref{fig:patching_b}, the patch dimension ($p$) defines the maximum number of connected units in both directions (i.e., visible $\rightarrow$ hidden and hidden $\rightarrow$ visible). 

\begin{figure}[hbtp]
\centering
	\includegraphics[width=0.46\textwidth]{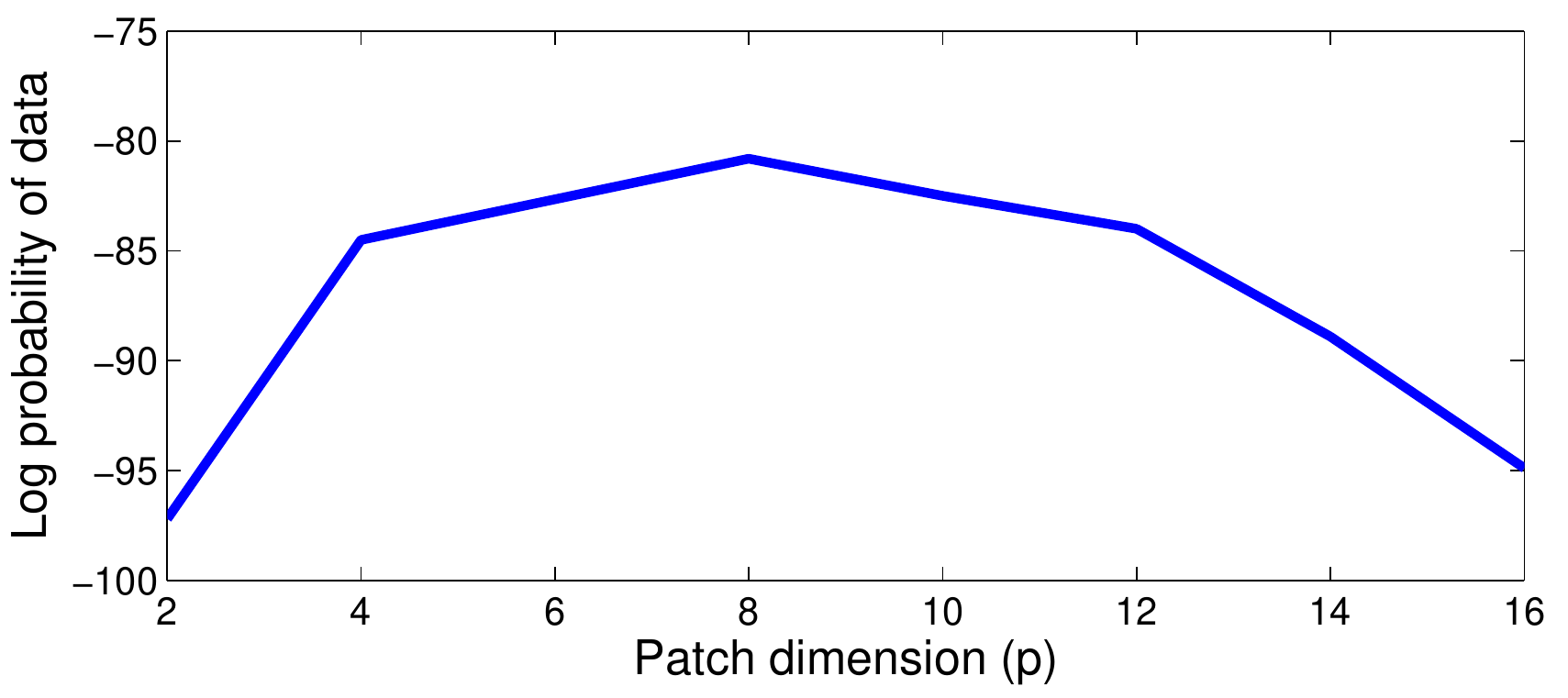}
	\caption{Generative performance versus sparsity.}
	\label{fig:sparse_quality}
\end{figure}

AIS measure versus patch dimension results are shown in Fig.~\ref{fig:sparse_quality}. For low $p$ values, lower log-probabilities were produced on account of less information captured by each patch. For large $p$ values, the log-probability is also lower on account of less number of hidden units (refer to Eq. \eqref{eq:patch}) in the network. Given the performance results of the model, for the generative RBM implementation an optimal patch size of 8$\times$8 was chosen, resulting in ($N$-$p$+1)$^2$ = (28-8+1)$^2$ = 441 hidden units.

\begin{figure*}[!ht]
\centering
	\begin{subfigure}{\textwidth}
	\centering
		\begin{subfigure}{0.68\textwidth}
			\centering
			\begin{subfigure}{1\textwidth}
				\centering
				\includegraphics[height=1.3cm]{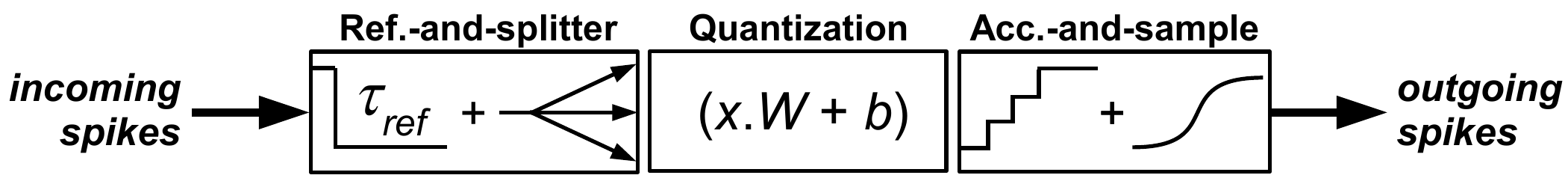}
				\caption{}
				\label{fig:tn_gen_a}
				\vspace{+4pt}
			\end{subfigure}\\
			\begin{subfigure}{1\textwidth}
				\centering
				\includegraphics[height=3.1cm]{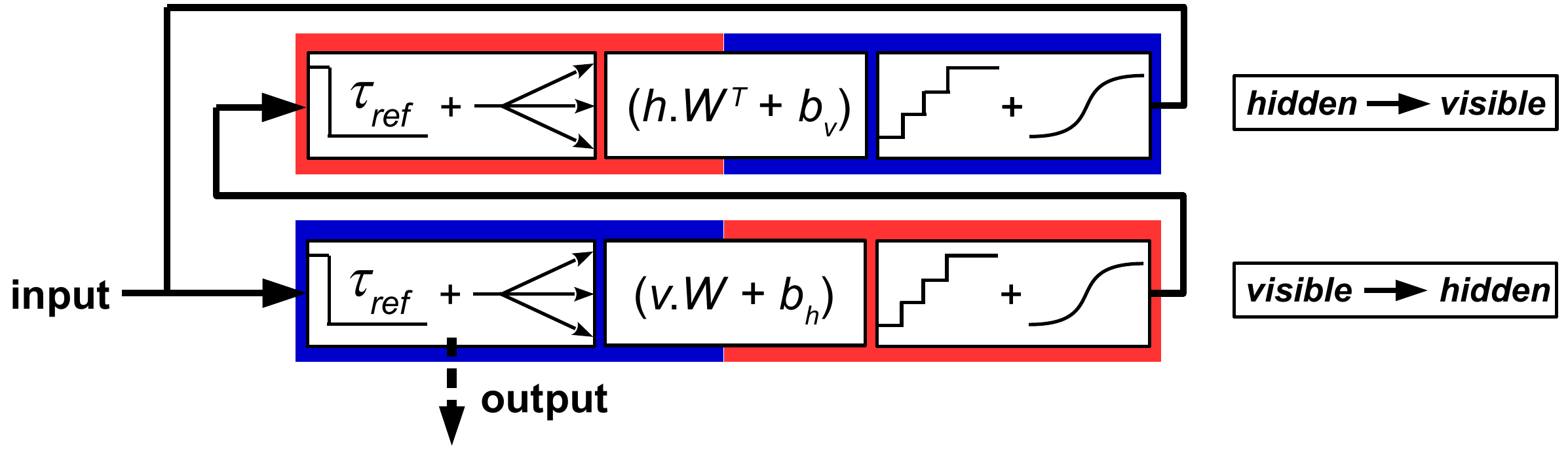}
				\caption{}
				\label{fig:tn_gen_b}
			\end{subfigure}
		\end{subfigure}%
		\begin{subfigure}{0.31\textwidth}
			\centering
			\includegraphics[height=5.1cm]{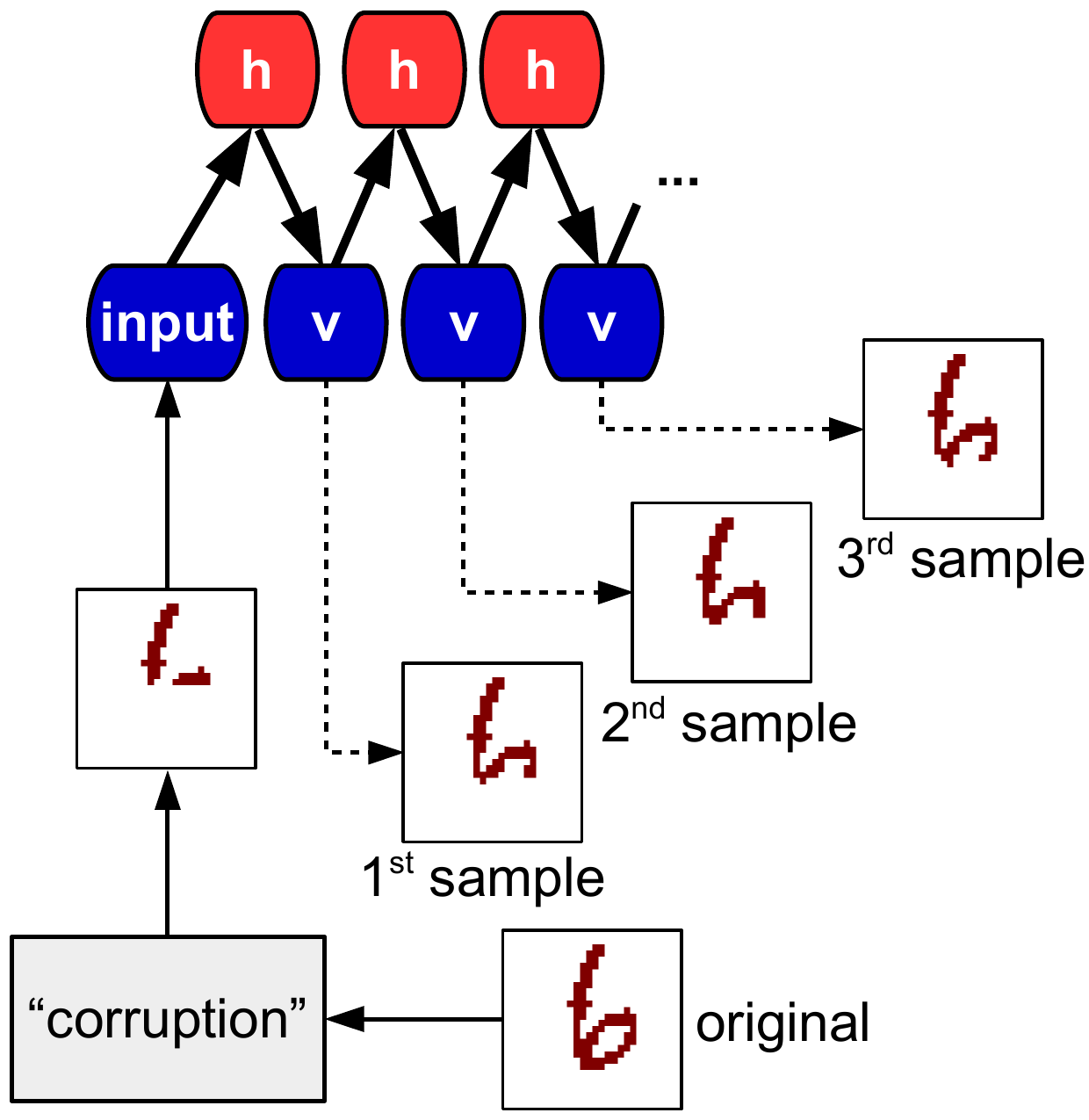}
			\caption{}
			\label{fig:tn_gen_c}
		\end{subfigure}
	\end{subfigure}
	\caption{TrueNorth RBM. (a) The 3-stage architecture used to distribute spike events (splitter), produce the desired membrane potential, and realize the sigmoid sampling. (b) The generative model structure, formed by combining two 3-stage blocks and including the feedback between layers. (c) Example of a pattern completion task where the digit ``6'' is incrementally reconstructed.}
	\label{fig:tn_generative}
\end{figure*}


\section{Generative RBM architecture on TrueNorth}
\label{sec:truenorth_rbm}
The generative RBM was mapped on TrueNorth by developing a modular 3-stage architecture, where each combination of these three stages represents the transition between RBM layers. The diversity of configurable parameters present in TrueNorth is critical to the realization, with particular neuron types, connectivity strategies, and reset modes in each stage. The physical constraints of TrueNorth -- particularly 256 axons and neurons per core, only 1 destination axon per neuron, and 4 distinct weights per neuron -- defined the design flow of the RBM. The architecture, composed of stages (1) refractory-and-splitter, (2) quantization and (3) accumulate-and-sample, is illustrated in Fig.~\ref{fig:tn_gen_a}.

The generative application implemented was a pattern completion task of a corrupted MNIST image. The signal flow in the TrueNorth RBM is illustrated in Fig.~\ref{fig:tn_gen_b}, where each row is a 3-stage module and the blue and red blocks represent information related to visible and hidden units, respectively. Note the second stage in each module contains both colors, since this is the transition between visible and hidden layers, i.e. where the arguments of $p(v|\textbf{h})$ and $p(h|\textbf{v})$ are computed. Finally, the data flow for the application is represented in Fig.~\ref{fig:tn_gen_c}. For this task, part of an image of the digit ``6'' (not used during training) was removed, and the figure shows the first 3 reconstructions based on the partial data.


\subsection{Stage 1a: Splitter}
\begin{figure*}[!ht]
		\centering
		\begin{subfigure}{0.32\textwidth}
			\centering
			\includegraphics[height=4.2cm]{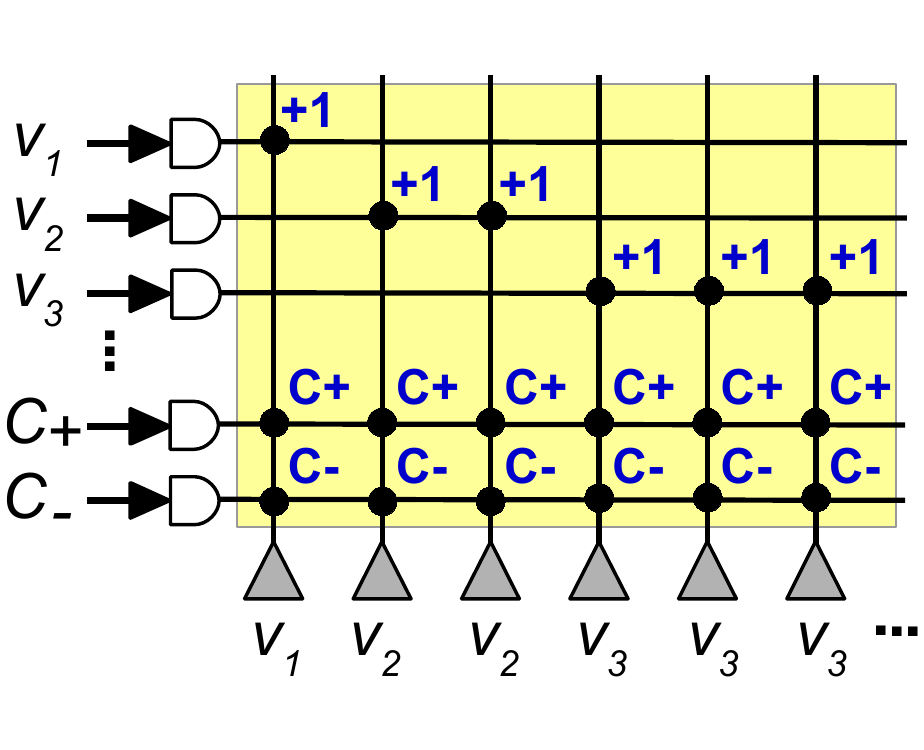}
			\caption{}
			\label{fig:3sa_1}		
		\end{subfigure}%
		\begin{subfigure}{0.32\textwidth}
			\centering
			\includegraphics[height=4.2cm]{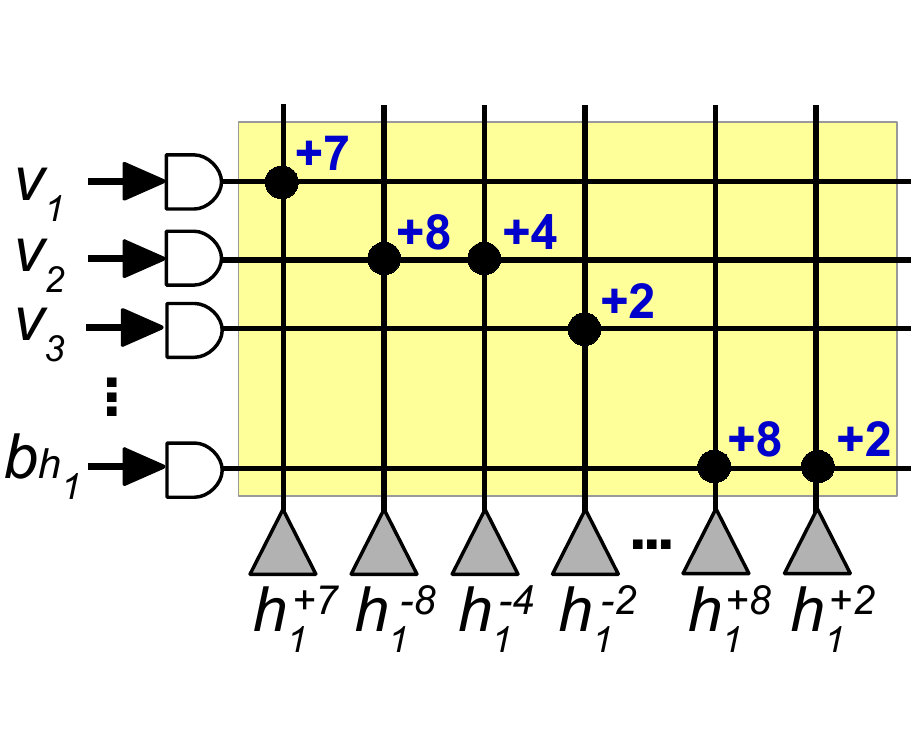}
			\caption{}
			\label{fig:3sa_2}		
		\end{subfigure}%
		\begin{subfigure}{0.32\textwidth}
			\centering
			\includegraphics[height=4cm]{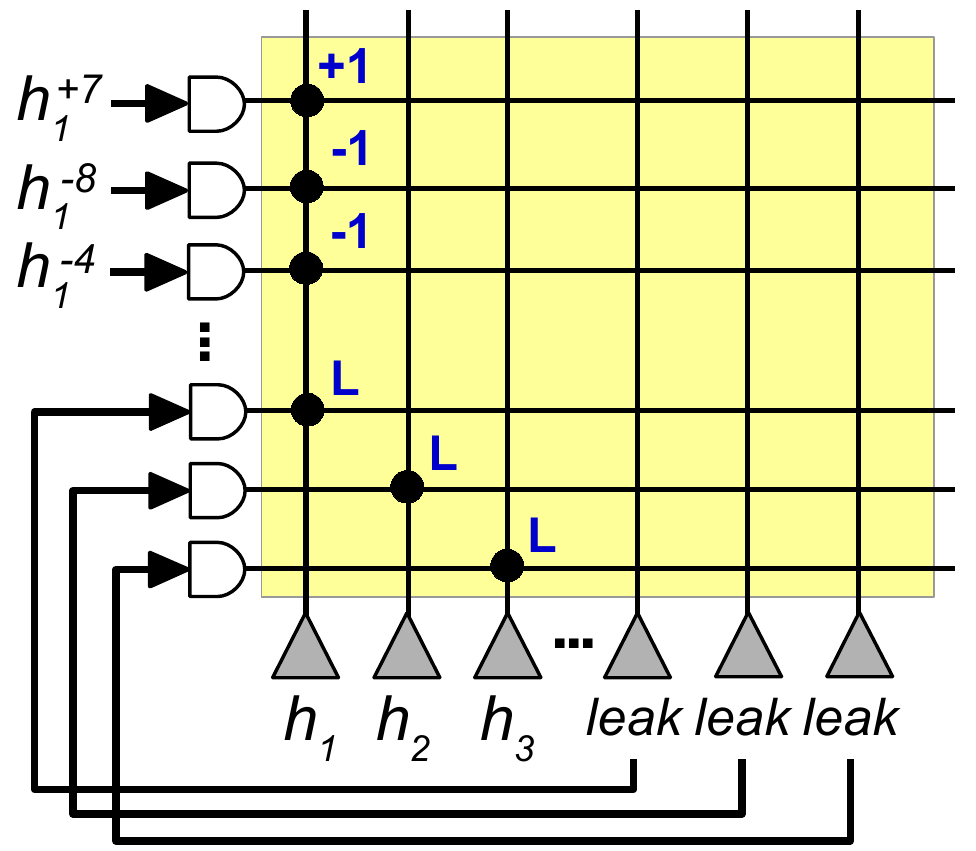}
			\caption{}
			\label{fig:3sa_3}		
		\end{subfigure}%
		\caption{Example of TrueNorth RBM stages: (a) Refractory-and-Splitter, (b) Quantization, and (c) Accumulate-and-Sample.}
		\label{fig:stages}
	\end{figure*}

Stage 1 serves a dual role in the system: (1) a splitter for input signals in the respective RBM layer and (2) a refractory effect of the neuron. Since each RBM unit in the visible/hidden layer is connected to multiple units in the hidden/visible layer, along with the fact that TrueNorth neurons present only one-to-one connections (i.e. each neuron can only target a single axon on the entire chip), a signal splitter is necessary to create the RBM's one-to-many connections. Therefore, stage 1 generates the required number of replicas of an RBM unit to be used in the quantization stage. Fig.~\ref{fig:3sa_1} illustrates a splitter core used for generating the necessary number of replicas of each of the visible units. The neurons are set to unit thresholds and all synaptic connections are of weight equal to +1, which will cause the neurons to spike whenever an axon event arrives. The refractory effect function of this stage and the two control signals ($C_+$ and $C_-$) are discussed later in Subsection~\ref{subsec:refractory}.

\subsection{Stage 2: Quantization}
\label{sub:quantization}
In TrueNorth, the weight of connections between axons and neurons can be configured with two constraints: the weights between axons connected to a given neuron are allowed to have only 4 different values; and each axon is configured as one of 4 types, reflecting on which of the 4 weights will be used for the connection between the axon and the respective neuron \cite{Cassidy13cognitivecomputing}. The first constraint limits the number of different possible weights, while the second limits the ``reutilization'' of axons between neurons. This is because an axon can be used amongst two neurons only if the weight stored in each neuron's memory position -- defined by the axon type -- is the desired synaptic weight for each of these connections. Several methods proposing the usage of low-precision weights and biases in artificial neural networks have been developed \cite{muller2015rounding,stromatias2015robustness}, however these methods target only discriminative models. As was observed in Figure \ref{fig:kl_div_scaling}, a large scaling factor (i.e. high precision) is critical for obtaining satisfactory generative performance in RBMs. Since the precision and diversity of weights and biases demanded by the generative RBM cannot be directly represented by the TrueNorth memory structure, a quantization stage is therefore necessary to realize the connectivity between RBM units.

The representation of individual RBM weights and biases was achieved by using a collection of neurons in stage 2, each with its own weight, which together can produce the desired membrane potential (i.e. the equivalent argument of $\sigma(x)$). For this, linear-reset, unit-threshold neurons are used \cite{Cassidy13cognitivecomputing}, and they operate by decrementing their membrane potential by 1 every time they spike, continuing to do so while the value is above zero. In this manner, the collective activity of many stage 2 neurons encodes the RBM weight/bias, while stage 3 will be used to accumulate the spikes from these many neurons into a single neuron.

The quantization of weights and biases is done by selecting a maximum accumulation time ($T_A$), which will be the largest value of membrane potential a stage 2 neuron can reach. In other words, every input spike into stage 2 axons will charge the membrane potential of each quantization neuron up to at most $T_A$, after which they will freely operate, with spiking activity guaranteed to cease in a maximum of $T_A$ ticks. Fig.~\ref{fig:3sa_2} exemplifies a stage 2 core with $T_A$=8 and visible units $v_1$, $v_2$, and $v_3$ connected to hidden unit $h_1$ with weights +7, -12, and -2, respectively. Since $T_A$ is a user-defined value, intuitively we would select the lowest value possible as to reduce the overall latency of the system. However, depending on the number of weights to be mapped and their specific values, attempting to use smaller values of $T_A$ will exceed the number of available neurons in a core. Note that the sign for the negative weights is actually positive, for stage 2 only takes into account the intensity (absolute value) of the connection between units, independent of being excitatory or inhibitory. The actual sign of the connection is taken care of in stage 3. Lastly, since bias values are independent of neuronal activity, these are realized by sending an external spike event to the bias axon ($b_{h_1}$=10 in Fig.~\ref{fig:3sa_2}) each time the sum of inputs to a given RBM unit neuron is to be computed.


\subsection{Stage 3: Accumulate-and-Sample}
Stage 3 is used to accumulate the activity of the quantization neurons into a single neuron, which will then be sampled (as described in Subsection \ref{subsec:quantization}). Prior to accumulation, the membrane potentials of the stage 3 neurons are initialized to zero. Then, during the first time window ($T_A$), stage 3 neurons accumulate spikes from stage 2 neurons to form a membrane potential equivalent to the argument of the logistic function. The neurons used in stage 3 have a non-resetting property to prevent clearing the membrane potential during the accumulation phase. This is followed by the time window $T_S$, during which the stochastic threshold and leak properties of the neuron are used for sampling from the logistic probability distribution. During this second time window, if the neuron's membrane potential surpasses the threshold, the neuron may spike multiple times since it is configured as non-resetting. For the spikes to correctly represent a sample from the logistic function, the refractory stage is necessary to register a maximum of 1 spike event per sampling window, and is described in the next subsection.

An example of a stage 3 core crossbar configuration is shown in Fig.~\ref{fig:3sa_3}, with the sign of the RBM weight/bias now included in the synaptic weights. Note the use of recurrent connections from additional neurons to realize the stochastic leak. These additional neurons are necessary because an internally generated stochastic leak (for example, in neurons $h_1$, $h_2$, and $h_3$) can only assume an absolute value of 1. Since our digital neural sampler implementation usually demands larger values of $L$, the ``leak'' neurons were created with threshold of 1 and internal stochastic leak sampled from a Bernoulli distribution with $p\approx 0.5$ (refer to Subsection \ref{subsec:quantization}). In this manner, there is approximately 50\% chance of these ``leak'' neurons spiking at each tick, thus generating a spike event to their respectively associated neuron ($h_1$, $h_2$, etc.), which can be connected with a user-defined synaptic weight of $L>1$.


\subsection{Stage 1b: Refractory effect}
\label{subsec:refractory}

As was discussed in Subsection \ref{subsec:quantization}, for the multiple spikes from the accumulate-and-sample stage to be converted to a single spike event -- which represents a sample from the logistic probability distribution --, stage 1 neurons were configured to produce a ``refractory effect''. What essentially occurs in stage 1 is a delayed propagation of the $spiked$ variable (in the digital neural sampler algorithm in Subsection \ref{subsec:algorithm}), whose value is dependent on spikes from the previous RBM layer's stage 3. This delayed response after $T_S$ has expired, therefore, results in a ``frame alignment'' (in the same 1 ms time step) of RBM unit samples to subsequent layers and guarantees precise operation of the generative RBM algorithm.

The refractory effect is obtained in TrueNorth by configuring stage 1 splitter neurons with a negative saturating membrane potential. The membrane potential of stage 1 neurons are initialized to the negative saturating value $C_-$, with $|C_-| \ge T_S$, at the start of the sampling phase of stage 3 in the other RBM layer. Every incoming spike in stage 1 will cause the membrane potential of its associated neuron to increase by 1. After $T_S$, the membrane potential of the stage 1 neurons are incremented by $C_+$ (= $|C_-|$), causing the neurons which received at least one spike to cross the threshold and simultaneously generate (``frame alignment'') a spike to the subsequent stage 2.


\section{Spike processing flow in TrueNorth RBM}
\label{sec:spike_flow}
\begin{figure}[hbtp]
	\centering
	\includegraphics[width=0.49\textwidth]{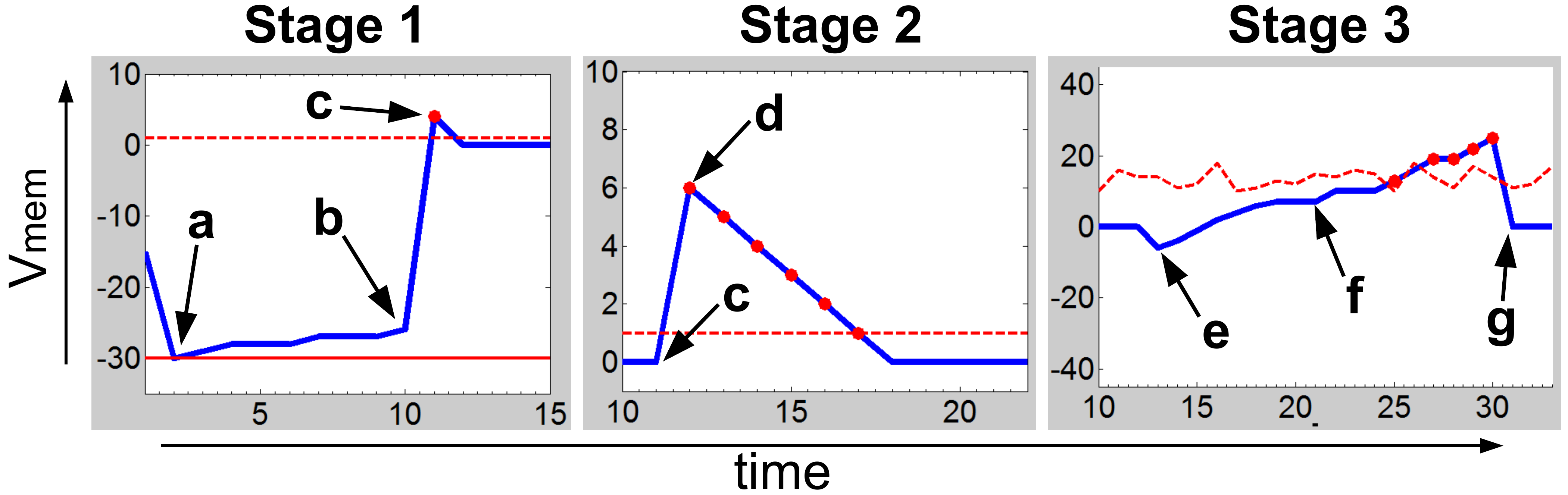}
	\caption{Example of spike processing flow. Stage 1 realizes two functions: refractory effect in (a) and (b), and splitter (spike distribution) in (c). Stage 2 quantizes the weights between RBM layers in (c) and (d), producing the desired membrane potential for stage 3 to sample from. Stage 3 accumulates the spikes from stage 2 linear-reset neurons between (e) and (f), and the sampling procedure is performed between (f) and (g). }
	\label{fig:3sa_flow}
\end{figure}

An example of the spiking activity flow between RBM layers is shown in Fig.~\ref{fig:3sa_flow}, where the following parameters were used: $T_S$=10, $T_A$=8, $C_-$=-30, stage 3 stochastic threshold ranging between 10 and 17 and stochastic leak of +3. In the example, the $x$-axis denotes time (in 1 ms ticks) and the $y$-axis denotes the value of the membrane potential ($V_{mem}$). The blue line is the neuron's membrane potential, the solid red line is the saturation level, the dashed red line is the threshold, and the red circles represent spike events.

The sequence of events (the letters) presented in Fig.~\ref{fig:3sa_flow} are detailed below:\\
(a) \textit{Time=2}. Stage 1 neurons are initialized to $C_-$, after which they begin accumulating spikes from stage 3 neurons of the other RBM layer. (b) \textit{Time=10}. After $T_S$, the $C_+$ signal is applied, and every neuron which captured at least one spike from stage 3 neurons crosses the threshold=1. (c) \textit{Time=11}. As $C_+$ is applied, spike events from stage 1 neurons are transmitted to stage 2 axons. In this example, the stage 2 neuron is charged to a membrane potential of 6. (d) \textit{Time=12-17}. The linear-reset stage 2 neurons continuously produce spike events to stage 3 axons until their membrane potentials return to zero. (e) \textit{Time=13}. At this moment, the stage 3 neurons begin accumulation for $T_A$ ticks. (f) \textit{Time=21}. After the stage 3 neurons have accumulated their membrane potentials to the desired values, the sampling phase begins. The stage 1 neurons of the other RBM layer are initialized to $C_-$; the stochastic threshold and leak come into effect at stage 3. (g) \textit{Time=31}. After $T_S$ ticks, the stage 3 neurons are reinitialized.

The example shows the complete sampling procedure of an RBM layer in the TrueNorth implementation. In stage 2, weights and biases are converted from membrane potential values to spikes, which are accumulated in stage 3 until the appropriate membrane potential is formed (i.e. has been grouped into a single neuron) at the start of the sampling phase. The two TrueNorth neurons in stage 3 comprise the coupled DTMCs used in the neural sampler. The stage 1 neurons produce the delayed spike response (``refractory effect'') in the subsequent RBM layer, which constitutes a sample from an RBM unit. Therefore, in this example, a new sample is produced in an RBM layer at every $(T_A + T_S + 2) = 20$ ticks; 2 additional ticks are necessary for control signals. The entire process of producing a new sample of the visible units -- the output of the generative RBM -- would then take $2 \times 20 = 40$ ticks (= 0.04 seconds).


\section{Design automation}
\label{sec:automation}
TrueNorth system configuration can be realized using the object-oriented \textit{Corelet} Language, which is an abstraction for representing the network of neurosynaptic cores \cite{amir2013cognitive}. The developed design automation procedure consists of creating systematic data structures, originating from the RBM weight and mask matrices, RBM biases, and user-defined parameters, which include: accumulation time (T$_A$); sampling time ($T_S$); data scaling factor ($s$); and sampler stochastic threshold and leak. Once these have been defined, the automation procedure produces an optimal configuration of cores which minimizes the number of axons and neurons used for the RBM realization. Three optimization strategies were created, where strategies 1.1 and 1.2 are mutually exclusive, yet they can be combined with strategies 2 and 3. Note that all considerations for hidden units are also valid for visible units.\\

\textbf{Strategy 1.1:}
The first strategy involves establishing the number of neurons required for mapping each RBM weight and bias. Without optimization in stage 2, the number of neurons $n_j$ used when quantizing the weight between the visible units observed by hidden unit $h_j$ can be computed by $n_j = \sum_{i} \lceil w_{ji}/T_A \rceil$. This direct method of mapping weights and biases does not take into account the fact that possibly many stage 2 neurons present low weights, which will cause them to complete spiking (during the accumulation phase) before neurons which represent higher values, such as weight $T_A$. Since the network must always go through $T_A$ ticks during the accumulation phase, it would be more efficient to try to connect a given neuron to as many possible axons, provided the total synaptic weight is guaranteed not to exceed $T_A$. In the limiting case, neurons which map weights -1 and +1 can have up to $T_A$ axons connected to them.

\begin{figure}[hbtp]
	\centering
	\begin{subfigure}[b]{0.23\textwidth}
		\centering
		\includegraphics[height=3.6cm]{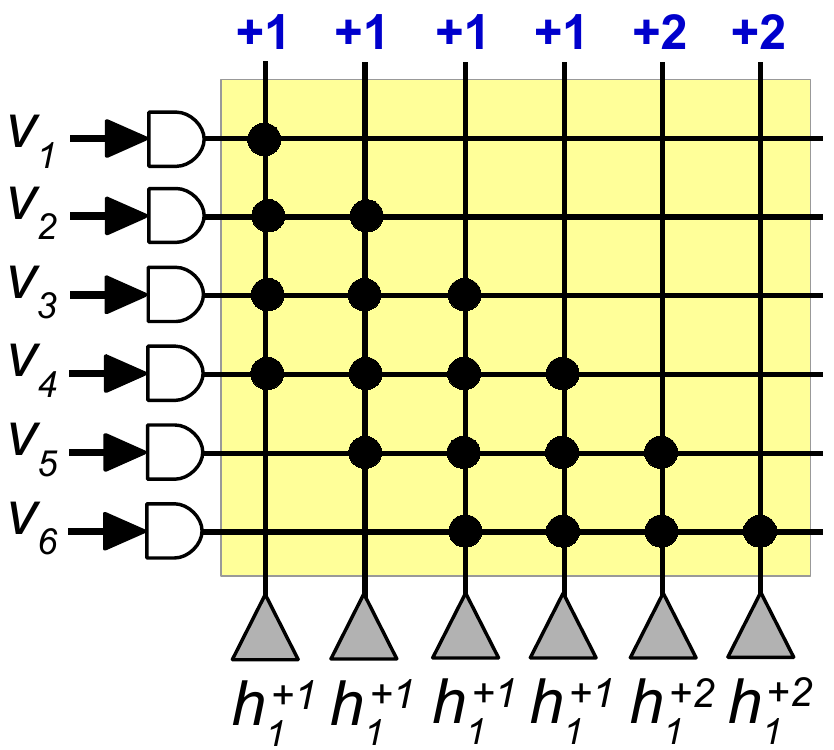}
		\caption{}
		\label{fig:opt_a}
	\end{subfigure}%
	\begin{subfigure}[b]{0.27\textwidth}
		\centering
		\includegraphics[height=3.6cm]{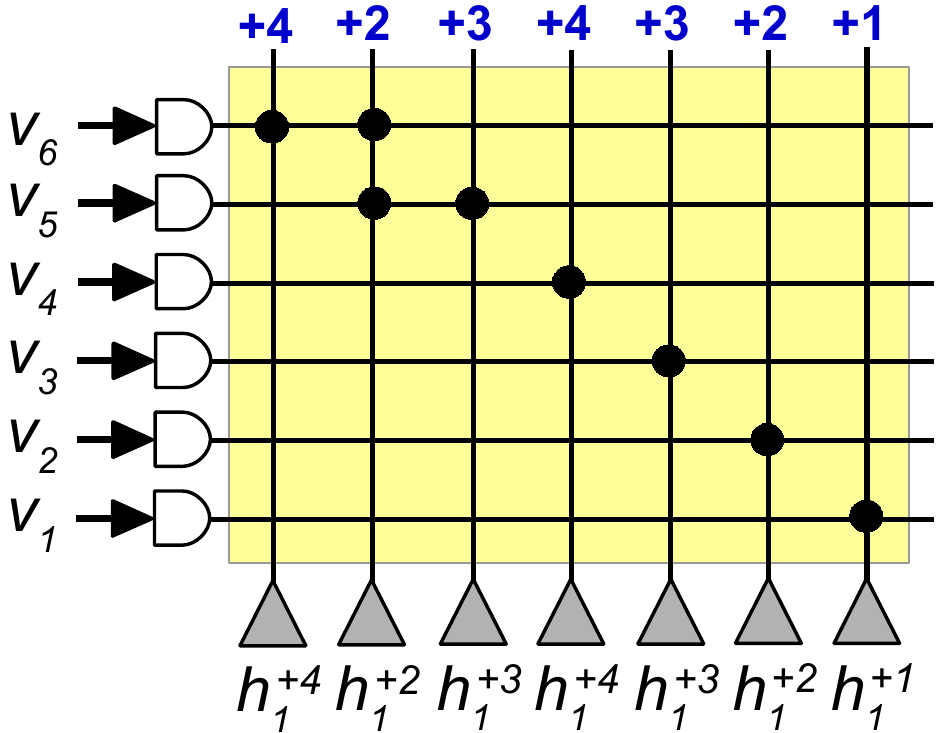}
		\caption{}
		\label{fig:opt_b}
	\end{subfigure}
	\caption{Strategy 1.1 examples of stage 2 quantization.}
\end{figure}

Though this first strategy benefits the core utilization considerably, better optimizations are possible. This is because the order in which the RBM weights are chosen to be mapped in stage 2 is defined by the user, yet different mapping sequences may utilize less cores. For example, suppose $T_A$=4 and the weights to be mapped are 1 through 6 for visible units $v_1$ through $v_6$, respectively. If we were to map them in this order, a total of 6 neurons would be used (Fig.~\ref{fig:opt_a}). On the other hand, if we were to map in the reverse order (6 through 1), a total of 7 neurons would be necessary (Fig.~\ref{fig:opt_b}). Therefore, the order of weight mapping affects the core utilization. Since the possible number of weight orderings to be analyzed is intractable, better results can be obtained by using strategy 1.2.\\

\begin{figure}[hbtp]
	\centering
	\includegraphics[width=0.48\textwidth]{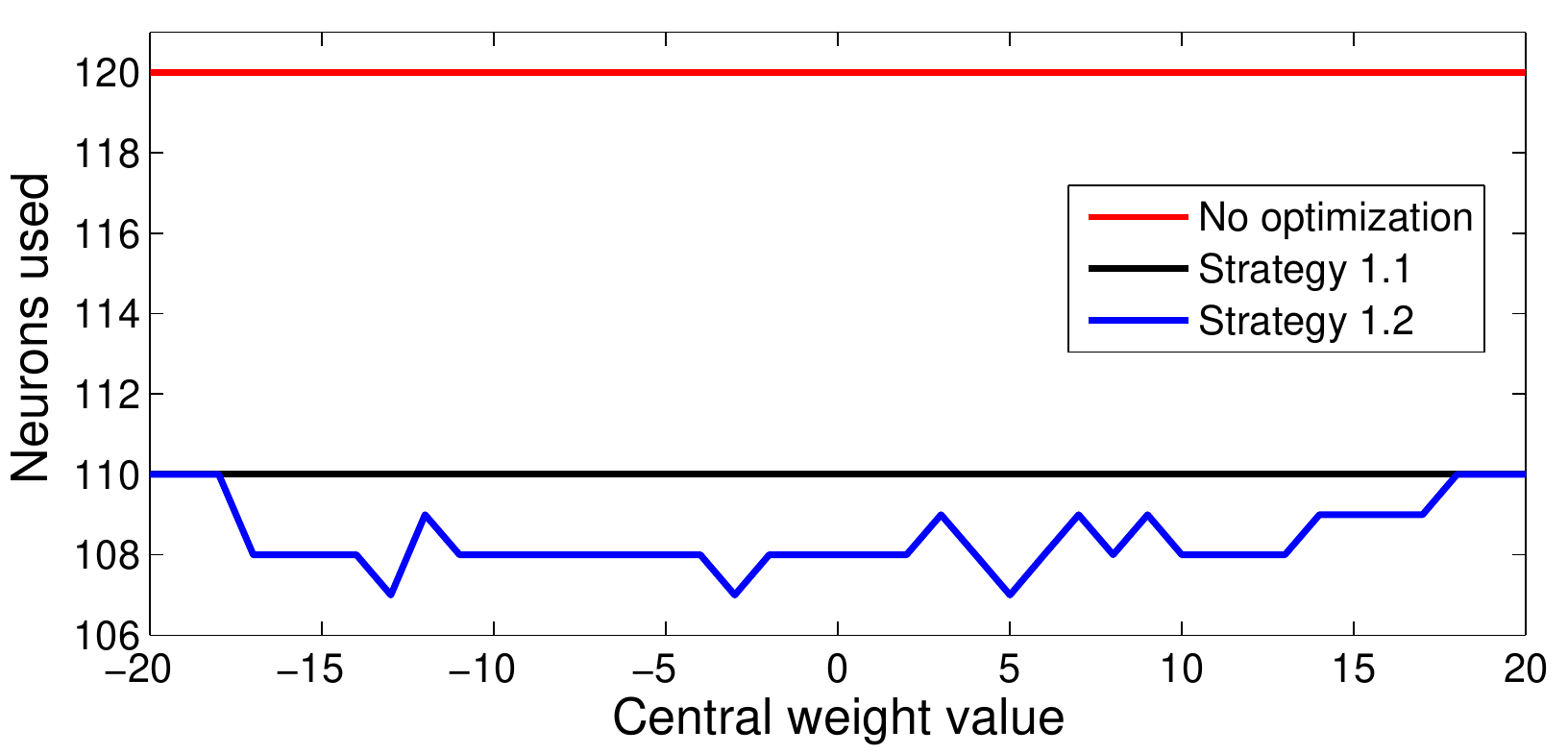}
	\caption{Optimization strategy comparison in terms of number of neurons used to map the desired weights.}
	\label{fig:opt_comp}
\end{figure}

\textbf{Strategy 1.2:}
In this strategy, the weights closest to a user-defined central weight value are mapped first. By sweeping through all possible central weights, an optimal value can be empirically obtained. Fig.~\ref{fig:opt_comp} shows the number of neurons used when mapping the weights -20 through 20 with $T_A=4$. The red line is the number of neurons (120) obtained with no optimization, while the black line is the number (110) when using sequential mapping with weight neuron ``reutilization`' (i.e. strategy 1.1). The blue line shows the results for the central weight method (strategy 1.2). The reduction from 110 to 107 neurons when using a central weight of 5, for example, is small (approximately 3\%), though more significant reductions are possible when this strategy is combined with strategies 2 and 3.

\textbf{Strategy 2:}
One final optimization can be performed in stage 2. When a hidden unit is mapped, the number of remaining neurons in the core may be enough to map additional hidden units. If this is the case, among the hidden units to be mapped, we select the one which has the most number of visible units in common with the hidden units previously mapped in the given core. This is because the patching scheme produces hidden units which may have some visible units in common and, thus, specific units are capable of sharing axons. This strategy is naturally also valid when mapping visible units.\\

\textbf{Strategy 3:}
Stages 1 and 3 can also be optimized via a greedy optimization method. For stage 1, the algorithm selects the unit which uses the most number of neurons (replicas for axons in stage 2), yet does not exceed the neuron limit in the core. If no unit can be mapped in the core, the algorithm creates a new one, until all units have been mapped. For stage 3, the algorithm does the same as for stage 1, now selecting the unit which uses the most number of axons (quantization neurons from stage 2) without exceeding the axon limit in the core.

\section{Results}
\label{sec:results}

For realizing the generative model -- the MNIST pattern completion task detailed in Section \ref{sec:truenorth_rbm} -- on TrueNorth, an RBM with 784 visible units and 441 hidden units (generated by using 8$\times$8 patches) was trained offline using the persistent Contrastive Divergence algorithm \cite{tieleman2008training}. The generative application demands a sampler with high fidelity with respect to the ideal sampler. To achieve this, the parameters were selected according to the criteria outlined in Section \ref{sec:quality}: scaling factor = 50, $T_S$=16, stochastic leak = 36, and stochastic threshold ranging from 186 to 697. The choice of the scaling factor directly impacts the RBM weight and bias magnitudes. To map these weights in stage 2, a trade-off is necessary between the accumulation time and the quantity of neurons and cores demanded by the application. Therefore, a compromise value of $T_A$=32 was selected for the mapping. With these parameters, a new RBM image is sampled at every 100 ticks (= 0.1 seconds).


\subsection{Resource utilization and power estimate}

Using the automation strategies outlined in Section~\ref{sec:automation}, the generative RBM was realized with 865 cores, representing 21\% of the total number of cores on TrueNorth. Table~\ref{tb:optimization} shows how applying the optimization strategies 1.2, 2 and 3 drastically reduced the core utilization. 

\begin{table}[h!]
\normalsize
  \begin{center}
    \begin{tabular}{| c || c | c |c|}
    \hline
     Case &  Strategies & \# of cores & Chip utilization \\
    \hline \hline
    1 & none & 2956 & 72.2\%\\
    2 & 1.1, 2, 3  & 906 & 22.1\%\\
    3 & 1.2, 2, 3 & 865 & 21.1\%\\
    \hline
    \end{tabular}
  \end{center}
  \caption{Core utilization results based on optimizations.}
	\label{tb:optimization}
\end{table}

In Figure \ref{fig:stages}, it was shown how each RBM unit is actually formed by 3 TrueNorth neurons (stochastic leak, stochastic threshold, and ``refractory effect''). However, the final implementation of the $784+441=1225$ RBM units consisted of 865 cores, with a total of 135k mapped TrueNorth neurons. This number is mainly due to the splitters and to the stages needed for weight and bias quantization, representing 82\% of the total neurons used. Virtually all of the remaining neurons were used for the control signals for the system operation, while the digital neural samplers -- which represent the RBM units \textit{per se} -- used up only 0.5\% of the mapped TrueNorth neurons. In practice, this results in a ratio of 110 TrueNorth neurons required to implement each RBM unit, and shows how generative models implemented on high dimensional datasets incur a considerable overhead due to the aforementioned hardware constraints. Nonetheless, given the network size (784 + 441 RBM units), image patch size ($p=8$), and accumulation ($T_A=32$) and sampling ($T_S=16$) times, we conservatively estimate a power consumption of 5 mW for the optimized TrueNorth generative RBM (case 3 in Table II). This results in an estimated 0.5 mJ of energy consumed to generate each MNIST image sample.

\subsection{Pattern completion outputs and metrics}

\begin{figure}[hbtp]
\centering
	\begin{subfigure}{0.32\textwidth}
		\centering
		\includegraphics[height=4.3cm]{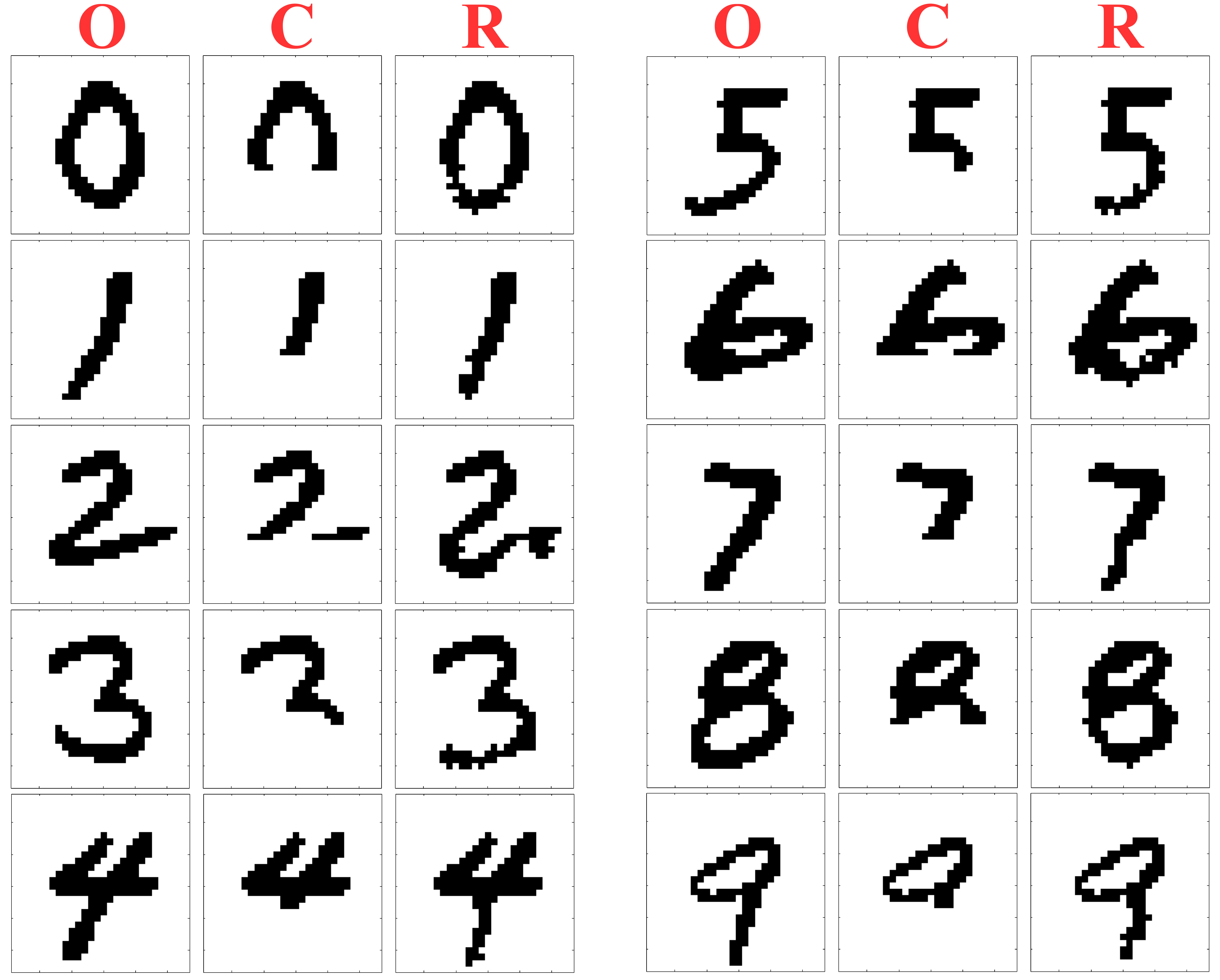}
		\caption{}
		\label{fig:results_good}
		\vspace{+10pt}
	\end{subfigure}%
	\begin{subfigure}{0.17\textwidth}
		\centering
		\includegraphics[height=4.3cm]{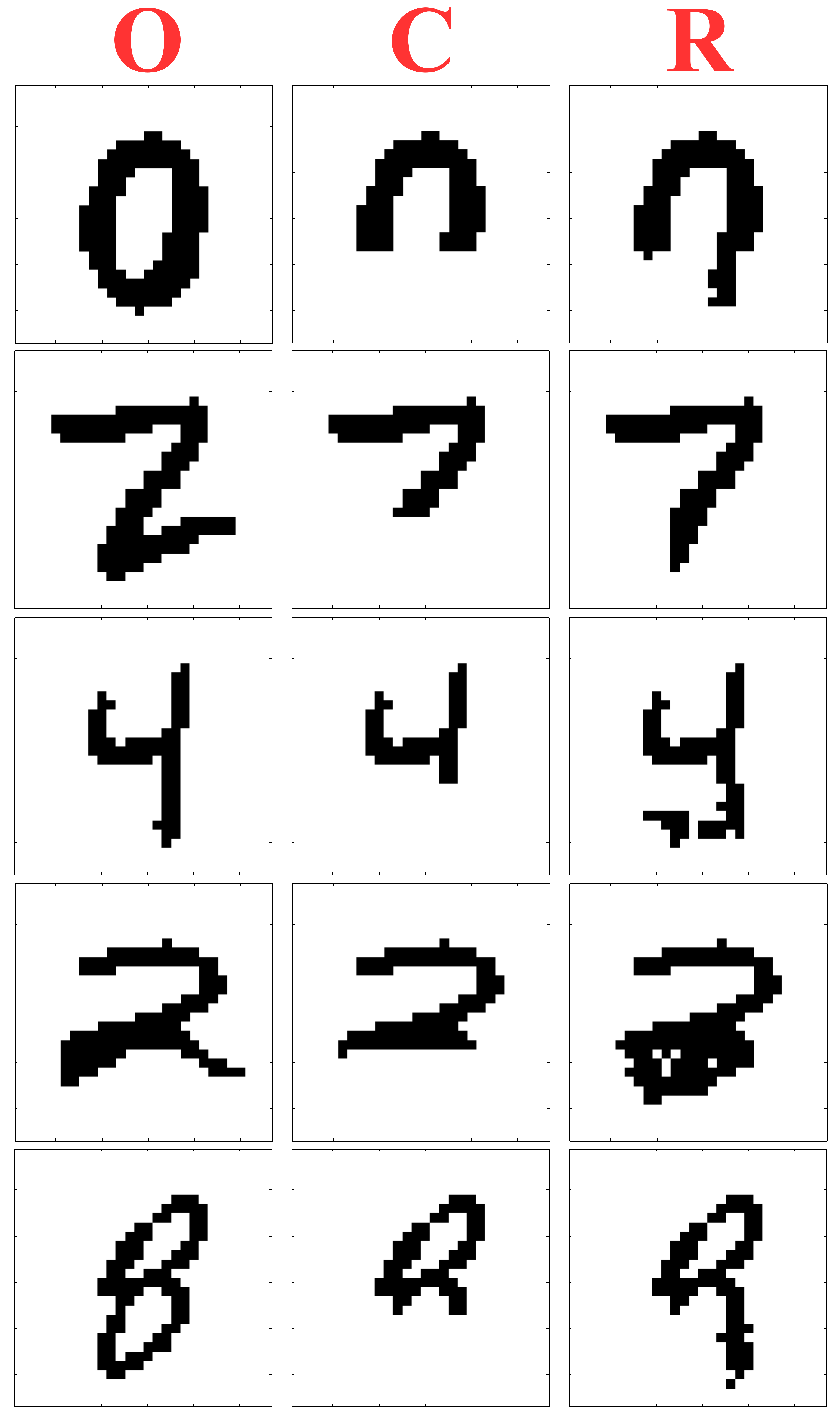}
		\caption{}
		\label{fig:results_bad}
		\vspace{+10pt}
	\end{subfigure}\\
	\begin{subfigure}{0.48\textwidth}
		\centering
		\includegraphics[width=1\textwidth]{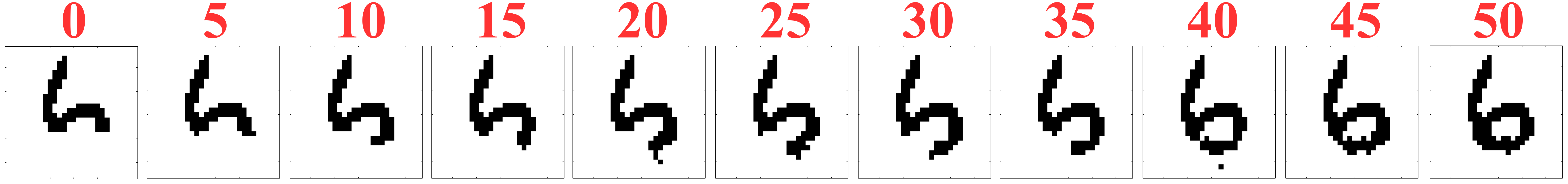}
		\caption{}
		\label{fig:results_long}
	\end{subfigure}
	\caption{TrueNorth pattern completion task outputs. Positive (a) and negative (b) reconstruction results. Reconstruction of the digit ``6'' in (c), with the sample number indicated above each image.}
\end{figure}

Example outputs of the pattern completion task are shown next. In Fig. \ref{fig:results_good}, one example output for each of the ten digits is presented: the first column is the original data (``O''), the middle column is the corrupted (``C'') image sent into the TrueNorth RBM, and the third column is the reconstructed (``R'') output after 50 RBM samples (= 5 seconds). These images were chosen to represent positive results, while Fig. \ref{fig:results_bad} shows images whose reconstruction was not ideal. Lastly, Fig. \ref{fig:results_long} illustrates a sequence of reconstructions for a corrupted image of the digit ``6''; the sample number is indicated above each image. A decent reconstruction sample could be obtained after about 4 seconds; however, with an earlier sample we could possibly confuse the ``6'' with a ``5''. 

\begin{figure}[hbtp]
\centering
\includegraphics[width=0.49\textwidth]{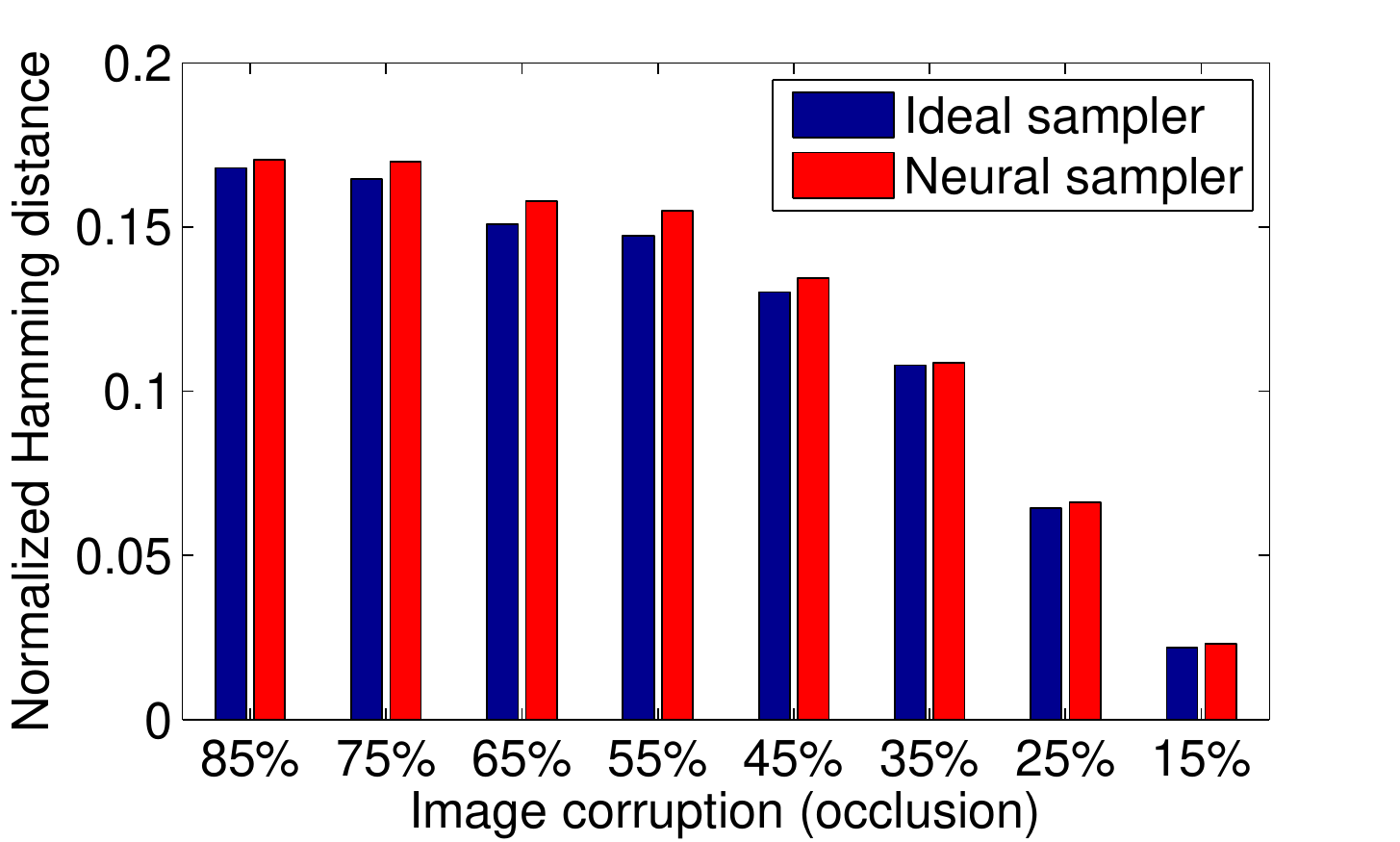}
\caption{Sampler generative performance analysis in terms of incorrectly generated pixels (Hamming distance).}
\label{fig:hamming}
\end{figure}

Depending on the percentage of image occlusion (``corruption''), the RBM may or may not be able to reconstruct a satisfactory representation of the original image. Therefore, we performed experiments with different image occlusion percentages and measured Hamming distance (HD) -- identical to the number of incorrectly reconstructed pixels in this case -- at the 50th RBM reconstruction sample for 1,000 test images. The mean value of the HDs was normalized according to the number of non-occluded pixels in the image. The results, illustrated in Fig. \ref{fig:hamming}, show that the reconstructive performance of the neural sampler nearly matches that of the ideal sigmoid sampler.

\begin{figure}[hbtp]
\centering
\includegraphics[width=0.49\textwidth]{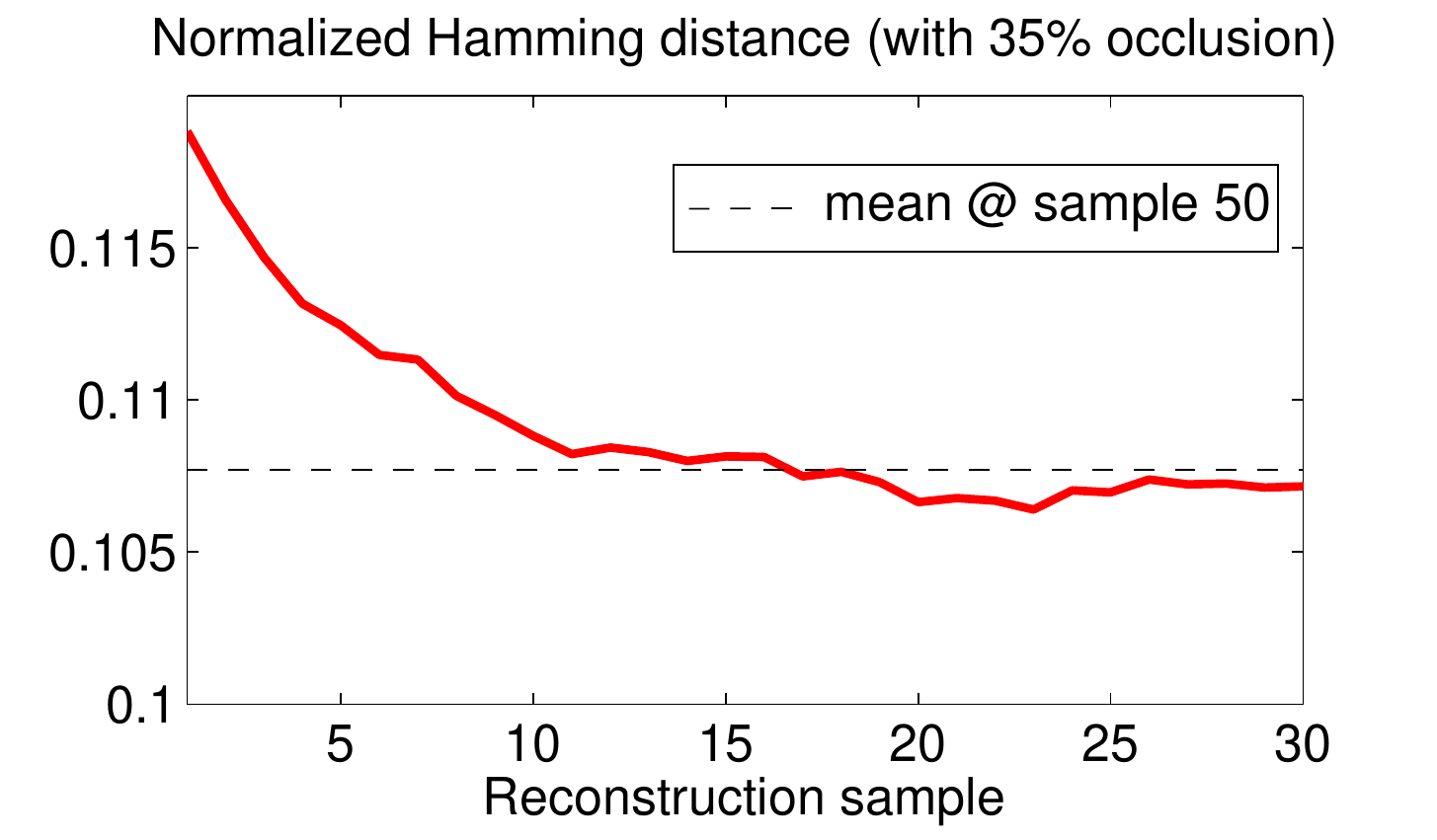}
\caption{Normalized Hamming distance during reconstruction of 35\% occluded image on the TrueNorth RBM.}
\label{fig:hamming2}
\end{figure}

Lastly, the mean HD for the TrueNorth RBM can be verified throughout the reconstruction process. In Fig. \ref{fig:hamming2}, convergence to the mean HD value for 35\% image occlusion (dashed black line) occurs after about 10 RBM reconstruction samples. This result is important to define the practical time expenditure demanded for the generative task of MNIST image reconstruction.

 \section{Conclusions and future work}

In this work, we have shown the first generative RBM implementation on neuromorphic hardware. For this, we followed a step-by-step procedure for producing the Gibbs sampling kernel -- the sigmoidal spiking probability -- using digital spiking neurons and for mapping the generative RBM algorithm onto a digital neuromorphic VLSI substrate. The neural sampler is an elegant solution as it uses bio-inspired dynamics to simultaneously incorporate the logistic function look-up and the comparison with a randomly generated number, which together represent a Gibbs sample. A discrete-time Markov chain (DTMC) analysis of the neural sampler was performed, resulting in a simplified method of obtaining the spiking probability without the need for long neuron behavior simulations. The generative performance of the neuromorphic adaptations were then verified using the Kullback-Leibler (KL) divergence and the Annealed Importance Sampling (AIS) algorithm. We also showed how mean squared error (MSE), along with the DTMC, can be used as an efficient method for obtaining insight into the sampler quality.

In the TrueNorth system, we followed a systematic development and implementation process of a modular architecture, which can be used for realizing generative RBMs and DBNs on a substrate of digital neurosynaptic cores. The 3-stage architecture and the design automation procedure provide a path towards automated neural network applications on brain-inspired processors for more complex inference tasks, such as natural image recognition and time series generation. The modular characteristic of the architecture naturally lends itself to implementations of deeper networks (DBNs). Also, the architecture of stages 1 and 2 with the associated design automation procedure can even be used to realize other neural networks which are defined by sparse weight matrices. We are currently working on new algorithms which incorporate more of the hardware constraints during the training phase.

The developed architecture avails many of the features present in neuromorphic systems. The spike flow of the 3-stage architecture developed for the TrueNorth RBM uses spikes for communication between cores, propagating information between RBM layers. The number of computations is also reduced in the neuromorphic scenario as only non-zero multiplications are performed (i.e. only when a spike occurs does data processing take place), which is contrary to what occurs traditionally for matrix multiplications in CPUs. Additionally, the sampler makes use of stochastic neural properties to produce an approximate sigmoidal firing probability, necessary for the RBM sampling procedure. Despite these positive features, information processing in the network is somewhat sequential (i.e. basically two stages are being used at each instant), which is mainly a result of the limited weight values per neuron in the present hardware. Inspired by the sampling methods proposed in \cite{oconnor2013real,neftci2013event}, we are currently developing paths towards algorithms on TrueNorth which incorporate the hardware constraints yet present a more continuous flow of spike processing for inference.


As a final note, research proposing RBMs and DBNs as solutions to applications of BCIs and EEG classification generally focuses on discriminative models \cite{wulsin2010semi,wulsin2011modeling,an2014deep}. However, BCIs could naturally benefit from generative models, targeting applications such as time series EEG or neural signal reconstruction for artificial limb control. An attractive feature of spike-based neuromorphic processors for spike-based neural interfaces would be the direct match between the event-driven data formats of the artificial and biological neuronal networks at the interface, potentially obviating the need for extra signal processing to convert between spiking and mean-rate representations, and possibly allowing to exploit the inherent temporal code of neuronal spike recordings or pulsed stimulation for further improvements in BCI performance.

\section*{Acknowledgment}

The authors would like to thank all the members of the Brain-Inspired Computing Group at the IBM Almaden Research Center and also the participants in the NSF Telluride Neuromorphic Cognition Engineering Workshop for their interaction and collaboration.

\ifCLASSOPTIONcaptionsoff
  \newpage
\fi



%

\bibliographystyle{ieeetr}
\bibliography{tbiocas_generative}  

%

%
%
%




\end{document}